\documentclass{article}

\usepackage{microtype}
\usepackage{graphicx}
\usepackage{subcaption}
\usepackage{tabularx}
\usepackage{booktabs} 
\usepackage[hyphens]{url}  
\usepackage{graphicx} 
\usepackage{natbib}  
\usepackage{caption} 
\usepackage{placeins}
\usepackage{algorithm}
\usepackage{algorithmic}
\usepackage{subcaption}
\usepackage{multirow}

\usepackage{hyperref}

\usepackage[accepted]{icml2026}

\usepackage{amsmath}
\usepackage{amssymb}
\usepackage{mathtools}
\usepackage{amsthm}

\usepackage[capitalize,noabbrev]{cleveref}

\theoremstyle{plain}

\theoremstyle{definition}

\theoremstyle{remark}

\usepackage[textsize=tiny]{todonotes}

\icmltitlerunning{Cross-Border Fulfillment Center Selection via Doubly Robust Estimation and Linear Programming}

\begin{document}

\twocolumn[
  \icmltitle{Cross-Border Fulfillment Center Selection via Doubly Robust Estimation and Linear Programming}

  \icmlsetsymbol{equal}{*}

  \begin{icmlauthorlist}
    \icmlauthor{Kriti Mahajan}{yyy}
    \icmlauthor{Patrick Bloebaum}{xxx}
  \end{icmlauthorlist}

  \icmlaffiliation{yyy}{Cross Border Tech \& Science, Amazon}
  \icmlaffiliation{xxx}{Agentic AI DevOps, Amazon}

  \icmlcorrespondingauthor{Kriti Mahajan}{kritimhj@amazon.com}

  \icmlkeywords{Machine Learning, ICML}

  \vskip 0.3in
]

\printAffiliationsAndNotice{}

\begin{abstract}
We introduce the cross-border (XB) fulfillment center (FC) Selection Model: a multi-stage optimization framework combining forecasting, doubly robust estimation, and linear programming to select optimal source FCs. For each source–destination shipping arc the objective is to choose a subset of source-country FCs that minimize delivery time and cost while maximizing profit and shipped units under a fixed domestic network. We evaluate the model on three production-scale international shipping corridors for a large e-commerce company using 12 months of shipment data, employing both aggregate offline replay and shipment-level counterfactual nearest-neighbor matching. Across all shipping corridors, the XB FC Selection Model achieves 8.5–51.1\% reductions in delivery time and 4.4–48.3\% reductions in logistics cost relative to human expert-defined FC selection policies, with up to 6.5 percentage point improvements in 5/5 target compliance (5 days, \$5 USD). Pareto analysis shows the model consistently dominates single-objective ranking strategies and unconstrained LP on the efficiency–reallocation frontier, with all results statistically significant.
\end{abstract}

\section{Introduction}


Cross-border (XB) e-commerce has grown rapidly, but unlike domestic operations, XB fulfillment must contend with customs regulations, volatile international freight conditions, and longer transit times while meeting rising customer expectations. Delivery speed and cost efficiency are critical to customer satisfaction and competitive performance, as long transit times and high delivery costs reduce conversion, retention, and repeat purchases \cite{deshpande2020} and inhibit cross-border growth 	\cite{li2022}. Therefore, effective cost control and lead-time reduction through well-designed logistics networks are essential for sustaining growth and competitiveness in XB e-commerce.

Given the above, a core challenge in XB order fulfillment is selecting source fulfillment centers (FCs)—origin warehouses that store, pick, pack, and dispatch orders for international customers. This FC selection is a key component of regionalization, which structures a fulfillment network into geographic regions so that customer orders are primarily served by nearby FCs, reducing distance, cost, and delivery time.  Unlike domestic fulfillment, where FCs can be newly sited or networks redesigned, XB operations typically operate within existing source-country infrastructure and transportation networks; the XB division cannot create new FCs or routes and must use the network as given. FC selection for each cross-border corridor therefore becomes a constrained subset selection problem. Moreover, the optimal FC subset for one source–destination corridor (e.g., US$>$Australia (AU)) may not generalize to another (e.g., US$>$Singapore (SG)) due to differing geographic, regulatory, and demand patterns. Selecting an inappropriate subset can consequently increase transit times and logistics costs thereby degrading customer experience. 

To address this challenge, we introduce the XB FC Selection Model, a data-driven framework integrating short-horizon demand forecasting, doubly robust estimation (DRE), and constrained linear programming to select source FC subsets for each XB corridor. Forecasting provides next-period operational metrics, DRE corrects selection bias in historical FC assignments and generates candidate FCs, and a multi-objective LP selects the final subset while controlling shipped-volume reallocation. The model minimizes delivery time and logistics cost while maximizing profit and shipped volume, implementing soft regionalization: orders are preferentially fulfilled by selected (in-region) FCs, while non-selected (out-of-region) FCs remain available for residual demand. Although expert-based FC selection performs competitively (Section 4), its manual process is difficult to scale across growing FC networks, corridors, and decision periods. The proposed ML framework by contrast enables scalable automated, forward-looking, multi-objective optimization beyond rank-based approaches. Existing ML-based regionalization methods for domestic fulfillment do not directly apply because XB operations must use a fixed source-country network without creating new FCs or routes. Offline evaluation on production-scale data shows 8.5–51.1\% lower delivery time, 4.4–48.3\% lower logistics cost, and up to 6.5 percentage-point higher 5/5 compliance versus expert policies. Ablation analysis (Sections 3–4) confirms that forecasting, DRE-based FC selection, and constrained LP optimization are all necessary for the model's observed Pareto dominance over comparable baselines.

\textbf{Related Work}: Our work sits at the intersection of supply chain optimization (SCO) and FC location choice, extending the limited literature on cross-border (XB) e-commerce logistics. E-commerce fulfillment has been extensively studied \cite{acimovic2015},\cite{jasin2019}. Prior work has studied large-scale e-commerce fulfillment optimization, including dynamic order fulfillment, transportation-network design, and domestic regionalization \cite{sinha2026}, \cite{oneill2023}, \cite{xu2009}, \cite{lara2023}, \cite{amazonscience2022},\cite{kassmann2006},\cite{allgor2023}. Cross-border logistics introduces additional regulatory and transportation complexity, and prior XB studies have considered overseas warehouse planning, logistics-mode choice, and network design \cite{cecil2024}, \cite{guan2021}.However, these approaches do not address the problem studied here: selecting a subset of existing source-country FCs for individual XB corridors under a fixed transportation network. Our approach combines causal estimation with forecasting and constrained optimization to address this production-scale FC subset-selection problem. The closest work in the XB FC selection literature is \cite{guan2021} proposes a game-theoretic framework for overseas FC siting; our paper differs in both theory and setup, using causal estimation and LP to select optimal FCs within the existing source-country network rather than creating new overseas facilities.

\begin{figure*}[t]
    \centering

    \begin{minipage}{0.48\textwidth}
        \centering
        \includegraphics[width=\linewidth]{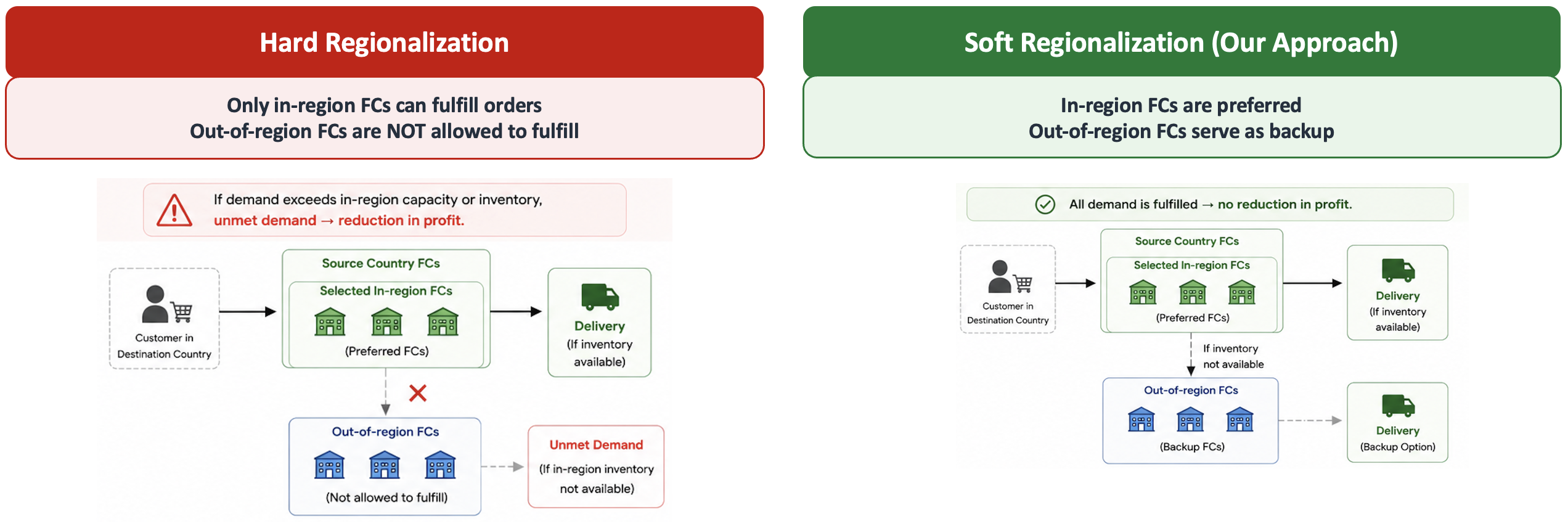}
        \caption{\small Hard vs. Soft Regionalization: Most businesses choose Soft Regionalization as it ensures zero demand loss while still capturing efficiency gains from better FC targeting}
        \label{fig:first}
    \end{minipage}
    \hfill
    \begin{minipage}{0.48\textwidth}
        \centering
        \includegraphics[width=\linewidth]{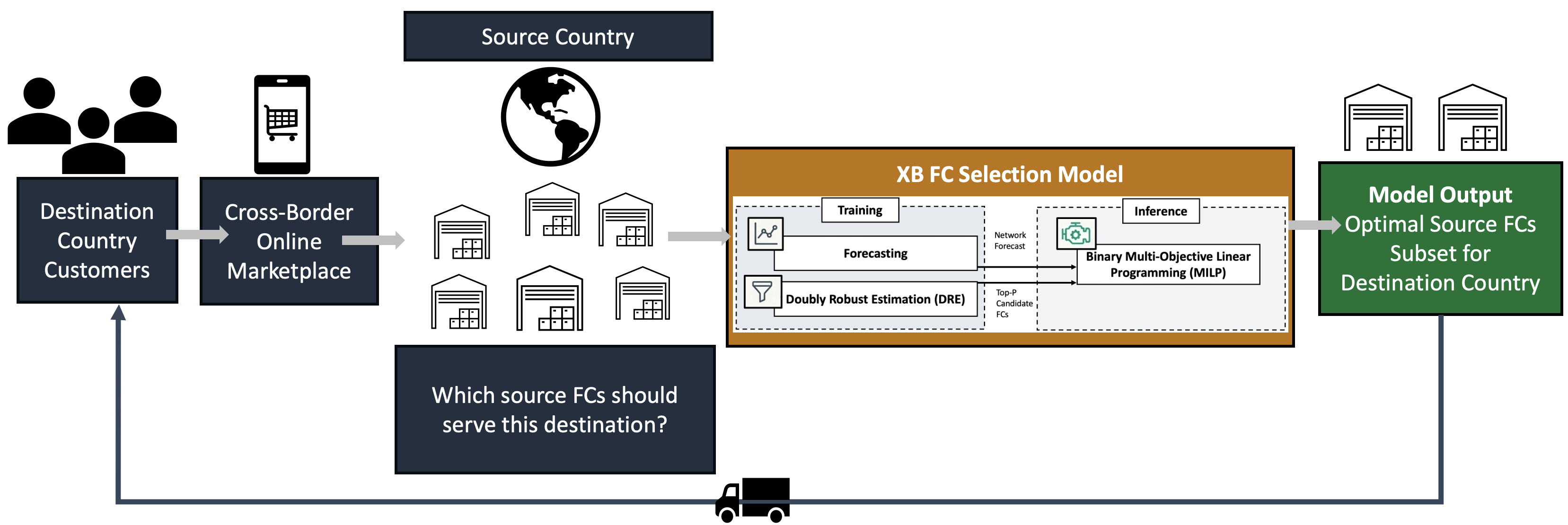}
        \caption{Cross-Border (XB) Fulfillment Center (FC) Selection Model}
        \label{fig:second}
    \end{minipage}

\end{figure*}

\section{XB FC Selection Model Architecture}

The XB FC Selection Model consists of three components: Short-horizon forecasting, Causal FC candidate generation, Constrained optimization. 


\textbf{2.1 Problem Set-Up and Notation} : Let $\mathcal{A} = \{(s,d)\}$ denote a set of source country $s$ to destination country $d$ shipping corridors; $\mathcal{W}_s$ denote the set of fulfillment centers located in source country $s$ ; $t \in \mathbb{N}$ index discrete decision periods (here days). Let $i \in \mathcal{W}_s$ index fulfillment centers. For each arc $a$, FC $i$, and time period $t$, we observe:  $T_{i,t,a} \in \{0,1\}$ - indicator if FC $i$ was active for $a$ on $t$, $Y_{i,t,a} \in \mathbb{R}$  - observed mean FC profit, and $X_{i,t,a} \in \mathbb{R}^d$ - the observed FC covariates.

Given that cross-border FC selection must balance multiple competing objectives, the observed covariates are:  $\tau_{i,t,a} :=$ Mean Delivery time from FC $i$ to destination $d$, $c_{i,t,a} :=$ Mean Shipping Cost from FC $i$ to destination $d$, $u_{i,t,a} :=$ Total Shipped Units from FC $i$ to destination $d$. $\{\tau, c, u, Y\}$ define the origin-destination (OD) flow values for FC $i$ on arc $a$ for each time period $t$. Note that delivery time is measured from source FC to destination country delivery pin code. Ship cost includes all costs across all types of transportation , taxes etc. Shipped Units is included to make sure that sufficient shipment volume is handled by the chosen FCs and is a proxy for product mix diversity. \footnote{This design is due to unavailability of FC capacity and XB inventory data, preventing direct inventory modeling. Based on stakeholder alignment, the LP assumes (1) all source-country inventory is XB-eligible and (2) XB inventory equals 2\% of total FC inventory, ensuring capacity for increased volumes in selected FCs. Shipped Units serves as a proxy for product-mix diversity.} For each arc $a = (s,d) \in \mathcal{A}$ and time $t$, the XB FC selection model's task is to select a subset of FCs in $s$ at time $t+1$ that will minimize delivery time and cost while maximizing shipped units and profit i.e. the model's task is to select $\mathcal{W}_{a,t} \subseteq \mathcal{W}_s$ of fulfillment centers. The FC selection has to be optimized in a forward looking fashion because due to technical constraints of the deployment platform, real (or near-real) time deployment of selected FCs is not possible. Thus, the model has to make a choice that is optimal for the next period $t+1$, given the information it has in the current period $t$.

The XB FC Selection Model is designed to achieve “5/5” through soft regionalization, optimizing FC-level profit in both the DRE and LP stages. Profit serves as a composite objective that balances cost efficiency and demand coverage: optimizing time or cost alone can concentrate volume in a few fast or cheap FCs, causing severe shipped-unit reallocation (up to 86\%, Section 4) and reducing in-region coverage. By contrast, profit penalizes such concentration because lower volume reduces total profit, acting as an endogenous regularizer that favors FC subsets maximizing in-region assignment while remaining cost- and time-efficient. DRE further benefits because profit captures each FC’s net causal contribution across operational dimensions.

\begin{algorithm}[t]
\caption{XB FC Selection Model}
\label{alg:main}
\begin{algorithmic}[1]
\REQUIRE Historical shipment data $\mathcal{D}$;
shipping corridors $\mathcal{A}$;
FC sets $\{\mathcal{W}_s\}$;
hyperparameters $(P,K_{\min},K_{\max},\alpha)$
\ENSURE Daily FC selection policy $\Pi(a,t)$

\FORALL{$a=(s,d)\in\mathcal{A}$}
    \STATE Aggregate shipment data into OD metrics
    $\mathcal{Z}_{i,t}$ for all $i\in\mathcal{W}_s$
    \STATE Forecast next-day metrics
    $\widehat{\mathcal{Z}}_{i,t+1}$ using short-horizon models
    \STATE Estimate per-FC treatment effects
    $\widehat{\mathrm{ATE}}_i$ via doubly robust learning
    \STATE $\mathcal{W}^{\mathrm{cand}}\gets$
    top-$P$ FCs with significant $\widehat{\mathrm{ATE}}_i$
    \STATE Solve binary LP over
    $\mathcal{W}^{\mathrm{cand}}$ using
    $\widehat{\mathcal{Z}}_{i,t+1}$
    \STATE $\Pi(a,t+1)\gets
    \{i\in\mathcal{W}^{\mathrm{cand}}:x_i=1\}$
\ENDFOR
\end{algorithmic}
\end{algorithm}



\textbf{2.2 Training Phase - Origin-Destination (OD) Forecasting:} Since FC selection at time $t$ affects outcomes at time $t+1$, we require forward-looking estimates. The forecasts are multivariate, jointly incorporating the temporal dynamics of the OD metrics $\tau$, $c$, and $u$. Specifically, for each OD metric $z \in \{\tau,c,u\}$, the target is the value of $z$ at time $t$, while the values of all three OD metrics at time $t-1$ are used jointly as predictors. Formally, for each arc $a$, we fit a multivariate forecasting model of the form $z_{a,t}
=
\beta_{z,\tau}\tau_{a,t-1}
+
\beta_{z,c}c_{a,t-1}
+
\beta_{z,u}u_{a,t-1}
+
\varepsilon_{a,t-1,z}$ where $z \in \{\tau,c,u\}$. 

\textbf{2.3 Training Phase - Candidate FC generation using Doubly Robust Estimation (DRE):} As mentioned previously, XB orders use shared domestic infrastructure, so FC selection is influenced by prior domestic-optimized policies, most notably domestic regionalization since 2022. Historical assignments $T_{i,t}$ are therefore non-random, reflecting heuristics, constraints, and capacity limits, making naive rankings by observed speed or profitability biased. We address this using Doubly Robust Estimation (DRE) to narrow the FC candidate set to 45, the business-defined maximum number of in-region FCs under soft regionalization. DRE estimates each FC’s causal impact on serving an arc despite biased and incomplete historical data, where only realized outcomes—not counterfactuals—are observed and past assignments are confounded by manual rules. This avoids systematically overvaluing frequently selected FCs and provides a more robust estimate of each FC’s true contribution. DRE is used to compute the Average Treatment Effect (ATE) for each origin FC $i$ as follows:  For each arc $a$, we model FC selection as a treatment where: $T_{i,t} = \mathbb{I}\{\text{FC $i$ selected}\}$, $Y_{i,t} = \pi_{i,t} (\text{Observed profit})$ and $X_{i,t-1} = \text{observed lagged confounders}$. The confounders are lagged at time t-1 to prevent bi-directionality and endogeneity. The observed lagged confounders are: \{\texttt{Distance\(_{t-1}\)}, \texttt{time\(_{t-1}\)}, \texttt{price\(_{t-1}\)}, \texttt{cost\(_{t-1}\)}, \texttt{shipped\_units\(_{t-1}\)}, \texttt{Regionalization} where Regionalization denotes the start of domestic regionalizations dummy indicator. We seek the ATE for each FC $i$: $\text{ATE}_i = \mathbb{E}\left[Y_{i,t}(1) - Y_{i,t}(0)\right]$. To generate the list of candidate FCs, for each arc we retain the top $P$ fulfillment centers with statistically significant positive ATEs:
$ \mathcal{W}_a^{\text{cand}} =
\left\{ i : \text{ATE}_i > 0,\; p\text{-value} < \alpha \right\}$.

\textbf{2.4 Inference Phase - Optimal top-k FC Selection via Multi-Objective Linear Programming:} We model the source FC selection problem as a binary linear program, optimizing multiple objectives: profit, shipped units, cost and time. We run the below optimization routine at each time period $t$ (for the purpose of this paper, the model is run daily) using the OD forecasted values for time $t+1$ to determine the optimal FC selection at time $t+1$.  The resulting problem is a binary linear program solved daily using a MILP solver. More formally: \textbf{2.4.1 Decision Variables:} For arc $a$ and time $t$, define: $x_i \in \{0,1\}, \quad i \in \mathcal{W}_a^{\text{cand}}$ where $x_i = 1$ indicates FC $i$ is selected. \textbf{2.4.2. Parameters}: Weights for profit, cost, and time, subject to $w_\pi + w_c + w_\tau = 1$, are tuned using Bayesian optimization (BO). The Bayesian optimization objective function maximizes a composite score combining delivery-time and cost reduction while penalizing profit reduction. For each candidate weight vector $(w_\pi,w_c,w_\tau)$, the resulting FC selections from the optimization formulation in Sections 2.4.3-2.4.4 are evaluated via aggregate offline replay (Section 3.3.1) on a held-out validation window preceding the evaluation period. Weights are re-estimated monthly using a rolling validation window of the most recent month's data. \textbf{2.4.3. Objective:} We optimize a weighted combination of normalized forecasted metrics with a small lexicographic bias:
\[
\max_x \sum_i x_i
\left(
w_\pi \tilde{\pi}_{i,t+1}
- w_c \tilde{c}_{i,t+1}
- w_\tau \tilde{\tau}_{i,t+1}
+ \varepsilon b_i
\right),
\]

where $\tilde{\cdot}$ denotes $z$-score normalization, and
$\varepsilon = 10^{-6} \ll 1$ is a small coefficient used
to enforce deterministic tie-breaking (e.g., based on the
hash or ID of FC $i$). \textbf{2.4.4. Constraints:}The optimization is subject to two constraints. The first limits the number of selected origin fulfillment
centers (FCs):
\begin{equation}
K_{\min} \leq \sum_i x_i \leq K_{\max},
\end{equation}
where $K_{\min}$ and $K_{\max}$ denote the lower and
upper bounds, respectively, on the number of selected
origin FCs. The second enforces minimum shipment coverage:
\begin{equation}
\sum_i x_i \hat{u}_{i,t+1}
\geq
\alpha \sum_i \hat{u}_{i,t+1},
\end{equation}
where $\alpha = 0.1$ specifies the minimum fraction
of total shipped units that must be covered by the
selected origin FCs. The number of selected FCs (top-$k$) varies by
shipping corridor (arc), since different destinations
may require different numbers of FCs to serve demand,
depending on local demand--supply dynamics.


\section{Experimental Methodology}

\textbf{3.1 Dataset Description}: We evaluate the XB FC Selection Model on three shipping corridors — US$>$SG (Singapore), US$>$AU (Australia), and US$>$DE (Germany) — spanning 12 months of production data. The underlying dataset consists of anonymized cross-border shipment records aggregated from order-level data to the FC $i$ × arc $a$ × day $t$ level. \textbf{OD Flow Dataset Construction:} For each FC $i$ × arc $a$ × day $t$ observation, we compute origin-destination (OD) flow metrics - mean delivery time (including in-transit and customs processing time), mean shipping cost (across all transport modes and duties), total shipped units, and mean FC profit. 

\textbf{3.2 Baselines:} We compare the XB FC Selection Model against nine baselines across four categories: \textbf{Rule-based rankings}: (1) \textit{Profit Rank}: top-$k$ FCs by prior-period profit; (2) \textit{Time Rank}: top-$k$ FCs by prior-period delivery speed. \textbf{ATE-augmented rankings}: (3) \textit{ATE-Profit Rank}: DRE candidate generation (top 45 FCs with significant positive ATEs), followed by profit-based top-$k$ selection; (4) \textit{ATE-Time Rank}: the same candidate generation followed by time-based top-$k$ selection. \textbf{LP-based methods}: (5) \textit{Vanilla LP}: top-$k$ selection using the multi-objective LP (Section 2.4) over all FCs; (6) \textit{Profit Rank + LP}: top 45 FCs by prior-period profit followed by LP-based top-$k$ selection; (7) \textit{Time Rank + LP}: top 45 FCs by prior-period delivery speed followed by LP-based top-$k$ selection. \textbf{Composite-score heuristics}: (8) \textit{Observed Score Top-$k$}: FCs are scored using previous-day observed metrics,
$s_i = w_{\pi}\hat{\pi}_i-w_c\hat{c}i-w{\tau}\hat{\tau}_i+w_u\hat{u}_i$,
where $\hat{\pi}_i,\hat{c}_i,\hat{\tau}i,\hat{u}i$ are within-day standardized (zero-mean, unit-variance) profit, shipping cost, transit time, and shipped volume, respectively, with fixed weights $(w{\pi},w_c,w{\tau},w_u)=(1,1,1,0.5)$, chosen a priori and not tuned on the evaluation period. The top-$k$ FCs are selected, followed by greedy volume-floor repair: high-volume FCs replace the lowest-scored selected FC only if selected volume strictly increases, until 10\% shipment-volume coverage is reached or no improving swap exists. (9) \textit{Forecasted Score Top-$k$}: identical to (8), but using forecasted $z$ values instead of lagged metrics. Since profit is not forecast by the existing predictive-model block, we fit a one-step linear regression, $\hat{\pi}^{f}_i=\beta_0+\beta_1\pi^{\text{lag-1}}_i$, on the pre-cutoff training split and apply it at inference; all other metrics use forecasts from the main model pipeline. All strategies are benchmarked against two reference policies: (i) \textbf{the Expert baseline} — FC selection policy defined by human supply chain experts using inventory stock based pairwise FC evaluation, distance filtering, and sort-point availability checks — and (ii) a \textbf{No Regionalization baseline} in which all source-country FCs are eligible for all shipping corridors; the Expert baseline is also measured against this policy.

\textbf{3.3 Evaluation Framework:} XB fulfillment's order-to-FC assignment system supports only one active FC selection policy per arc, precluding within-arc randomization and conventional A/B testing. Because alternative policies require a full-volume cutover rather than controlled diversion, strong offline performance guarantees are needed before any online pilot. We therefore use a two-stage evaluation protocol: (i) aggregate offline replay on historical data, followed by (ii) shadow deployment, where candidate FC selection strategies $f$ make daily FC selections using live data in parallel with the no-regionalization baseline, without affecting live routing. We use two offline evaluation frameworks: \textbf{3.3.1 Aggregate Offline Replay:} Each strategy $f$ from Section 3.2, including the XB FC Selection model, makes daily FC selections using only information available at time $t$. We simulate the network assuming only $f$'s selected FCs are active and compute outcomes from production data at $t+1$. Thus, if $f$ selects 10 of 100 FCs on a corridor, we characterize the resulting network using the aggregate mean time, cost, and shipped volume of those 10 FCs. Performance is aggregated monthly and reported relative to baseline as $\%\text{ change} = (\text{Baseline Value}-\text{Strategy Value})/\text{Baseline Value}\times100$. We report Time (days; larger reduction is better), Cost (thousands USD; larger reduction is better), and Shipped Units (thousands; smaller reduction is better) for both pre- and post-shadow deployment windows to assess temporal stability. Under soft regionalization, all demand is fulfilled; thus, ``shipped unit reduction'' reflects volume handled by non-optimal (out-of-region) FCs, not unserved demand.\textbf{3.3.2 Counterfactual Matching (only for post-shadow deployment)}: Offline evaluation cannot observe what would have occurred if a different FC had fulfilled an XB order. We therefore partition shipments into \emph{optimal} FC orders (fulfilled by an FC in strategy $f$'s recommended set) and \emph{non-optimal} FC orders (fulfilled by an FC outside the set), and construct shipment-level counterfactuals using two complementary matching approaches. \textbf{Same-ASIN Matching:} For each non-optimal shipment, we match historical shipments of the same ASIN fulfilled by an optimal FC and use their realized delivery time and shipping cost as counterfactuals. This is highly sparse, however, with fewer than 1\% of daily demand having the same ASIN assigned to both optimal and non-optimal FCs.\textbf{Nearest-Neighbor Matching:} We therefore use nearest-neighbor matching as a complementary approach for at-scale counterfactual evaluation. For each non-optimal order $i$ with standardized package dimensions $\mathbf{x}i\in\mathbb{R}^3$ and GL $g_i$, we find the nearest optimal order
$j^*=\arg\min{j\in\mathcal{O}^{\mathrm{in}}_g}|\mathbf{x}_i-\mathbf{x}_j|_2$
using a GL-stratified KD-Tree. The matched order's realized delivery time and shipping cost provide counterfactual estimates for $i$. These estimates are then used to compute Mean Delivery Time, Mean Shipping Cost, and \% Meeting the 5/5 Target (delivery $\leq 5$ days and cost $\leq \$5$ USD) for each strategy.

\textbf{3.4 Implementation Details}: AutoML fits the forecasting functions, with each variable/arc trained through a period ending one month before forecasting period $t$ to prevent look-ahead bias. OD forecasting and DRE-based candidate generation run in parallel. DRE uses EconML's DRLearner, with propensity and outcome models automatically selected from linear and gradient-boosted forest models; the final CATE model is estimated by OLS. LP optimization uses CBC via PuLP. The selected FC count ($topk$) is 11/6 for US$>$SG (pre/post-shadow), 8 for US$>$DE, and 6 for US$>$AU.

\section{Results}

\begin{table*}[t]
\centering
\tiny
\setlength{\tabcolsep}{3pt}
\caption{FC Selection Strategies Comparison Across Regions -- Historical Performance Relative to Expert FC Selection (\% Values).}
\label{tab:fc_comparison_all_regions}

\begin{tabular}{l|rrr|rrr|rrr|rrr|rrr|rrr}
\toprule

& \multicolumn{6}{c|}{US$>$SG}
& \multicolumn{6}{c|}{US$>$DE}
& \multicolumn{6}{c}{US$>$AU} \\

& \multicolumn{3}{c|}{Pre-Deployment}
& \multicolumn{3}{c|}{Post-Deployment}
& \multicolumn{3}{c|}{Pre-Deployment}
& \multicolumn{3}{c|}{Post-Deployment}
& \multicolumn{3}{c|}{Pre-Deployment}
& \multicolumn{3}{c}{Post-Deployment} \\

\cmidrule(lr){2-4}
\cmidrule(lr){5-7}
\cmidrule(lr){8-10}
\cmidrule(lr){11-13}
\cmidrule(lr){14-16}
\cmidrule(lr){17-19}

Strategy
& Time & Cost & Units
& Time & Cost & Units
& Time & Cost & Units
& Time & Cost & Units
& Time & Cost & Units
& Time & Cost & Units \\

\midrule

Profit Rank
& 0.092 & -0.104 & -0.050
& 0.080 & -0.132 & -0.120
& 0.405 & -0.230 & -0.273
& 0.110 & -0.197 & -0.206
& 0.316 & -0.067 & -0.068
& 0.193 & -0.043 & -0.040 \\

Time Rank
& -0.477 & -0.452 & -0.463
& -0.113 & -0.627 & -0.628
& -0.157 & -0.726 & -0.735
& -0.640 & -0.850 & -0.865
& 0.378 & -0.851 & -0.862
& 0.476 & -0.857 & -0.869 \\

ATE-Profit Rank
& -0.015 & -0.111 & -0.048
& 0.013 & -0.131 & -0.126
& 0.274 & -0.460 & -0.471
& 0.010 & -0.436 & -0.439
& 0.626 & -0.375 & -0.374
& 0.419 & -0.316 & -0.292 \\

ATE-Time Rank
& -0.412 & -0.162 & -0.142
& -0.274 & -0.414 & -0.421
& -0.379 & -0.691 & -0.667
& -0.627 & -0.706 & -0.690
& -0.187 & -0.582 & -0.589
& -0.094 & -0.596 & -0.584 \\

LP
& -0.453 & 0.001 & 0.070
& -0.187 & -0.056 & 0.040
& -0.223 & -0.822 & -0.834
& -0.613 & -0.883 & -0.893
& 0.397 & -0.939 & -0.940
& 0.382 & -0.938 & -0.942 \\

XB FC Model (ATE-LP)
& -0.418 & -0.044 & 0.025
& -0.244 & -0.086 & -0.077
& -0.271 & -0.459 & -0.489
& -0.511 & -0.483 & -0.485
& -0.085 & -0.364 & -0.359
& -0.133 & -0.326 & -0.301 \\

Profit Rank + LP
& -0.486 & -0.758 & -0.789
& -0.130 & -0.834 & -0.786
& -0.263 & -0.612 & -0.625
& -0.615 & -0.671 & -0.668
& -0.057 & -0.790 & -0.804
& 0.026 & -0.784 & -0.803 \\

Time Rank + LP
& -0.574 & -0.067 & 0.002
& -0.321 & -0.119 & -0.021
& -0.323 & -0.765 & -0.774
& -0.704 & -0.823 & -0.831
& -0.383 & -0.820 & -0.833
& -0.367 & -0.813 & -0.826 \\

Observed Score Top-k
& -0.603 & -0.305 & -0.238
& -0.203 & -0.282 & -0.176
& -0.178 & -0.489 & -0.479
& -0.344 & -0.385 & -0.373
& 0.148 & -0.135 & -0.127
& 0.219 & -0.152 & -0.129 \\

Forecasted Score Top-k
& -0.612 & -0.370 & -0.315
& -0.196 & -0.280 & -0.188
& -0.181 & -0.487 & -0.499
& -0.338 & -0.386 & -0.383
& 0.190 & -0.128 & -0.126
& 0.175 & -0.101 & -0.090 \\

\bottomrule
\end{tabular}
\end{table*}

\begin{table}[htbp]
\centering
\small
\caption{Shipment-Level Counterfactual Evaluation: Same Product and Nearest Neighbor from Optimal vs.\ Non-Optimal FC}
\resizebox{\linewidth}{!}{
\begin{tabular}{llcccccccccccc}
\toprule
& & \multicolumn{4}{c}{SG} & \multicolumn{4}{c}{DE} & \multicolumn{4}{c}{AU} \\
\cmidrule(lr){3-6} \cmidrule(lr){7-10} \cmidrule(lr){11-14}
Evaluation & Metric 
& Actual & Expert & LP & XB FC 
& Actual & Expert & LP & XB FC 
& Actual & Expert & LP & XB FC \\
\midrule

\multirow{4}{*}{Same Product}
& Mean Time (Days) 
& 3.21 & 2.74 & 1.36 & 1.18 
& 5.42 & 5.87 & 3.21 & 2.76 
& 4.68 & 2.31 & 2.47 & 1.84 \\

& Mean Cost (USD) 
& 8.47 & 8.31 & 7.14 & 6.52 
& 13.76 & 13.21 & 11.84 & 10.27 
& 14.63 & 12.46 & 14.17 & 11.92 \\

& \% Meeting 5/5 
& 54.72 & 47.35 & 51.86 & 59.41 
& 11.64 & 6.27 & 9.18 & 12.83 
& 6.02 & 6.47 & 4.16 & 8.21 \\

& Sample Size (N)
& 1,842 & 1,842 & 1,842 & 1,842
& 1,176 & 1,176 & 1,176 & 1,176
& 963 & 963 & 963 & 963 \\

\midrule

\multirow{4}{*}{Nearest Neighbor}
& Mean Time (Days) 
& 1.38 & 1.23 & 0.72 & 0.70 
& 3.03 & 3.23 & 2.00 & 1.85 
& 2.53 & 0.93 & 1.14 & 1.13 \\

& Mean Cost (USD) 
& 5.86 & 5.87 & 5.77 & 5.55 
& 11.08 & 10.68 & 10.03 & 9.36 
& 12.02 & 11.13 & 12.28 & 10.99 \\

& \% Meeting 5/5 
& 41.48 & 38.10 & 40.64 & 44.55 
& 6.83 & 3.18 & 5.61 & 6.85 
& 3.88 & 4.66 & 2.46 & 3.57 \\

& Sample Size (N)
& 94,216 & 94,216 & 94,216 & 94,216
& 87,543 & 87,543 & 87,543 & 87,543
& 81,769 & 81,769 & 81,769 & 81,769 \\

\bottomrule
\end{tabular}
}
\label{tab:counterfactual_summary}
\end{table}

\begin{figure*}[t]
    \centering
    \begin{subfigure}[b]{1\textwidth}
        \centering
        \includegraphics[width=0.7\textwidth]{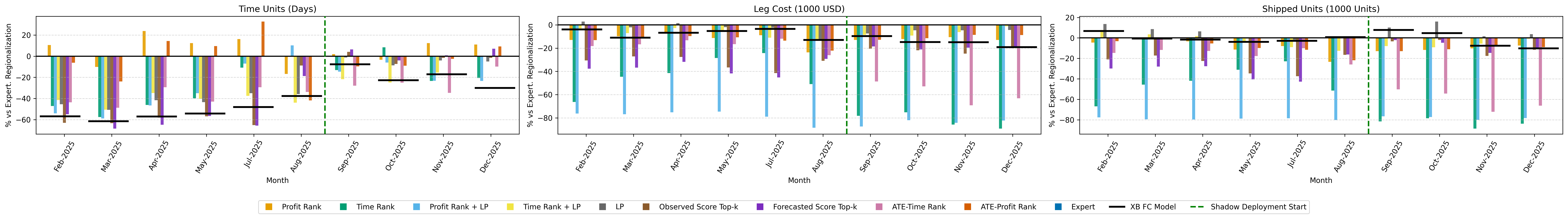}
        \caption{US$>$SG}
        \label{fig:graph1}
    \end{subfigure}
    
    
    \begin{subfigure}[b]{1\textwidth}
        \centering
        \includegraphics[width=0.7\textwidth]{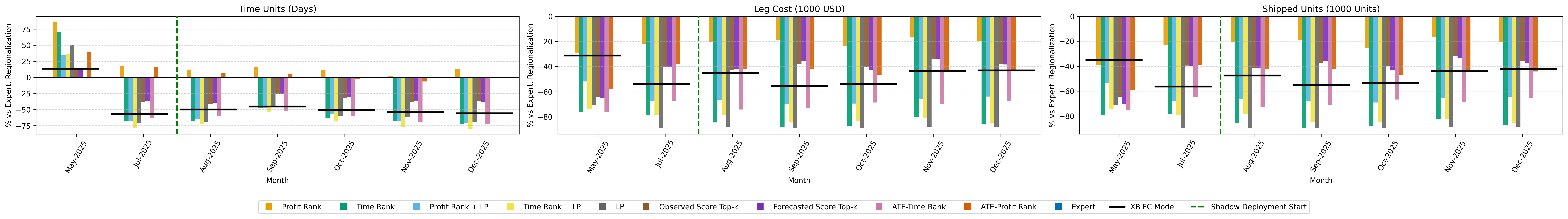}
        \caption{US$>$DE}
        \label{fig:graph2}
    \end{subfigure}
    
    
    \begin{subfigure}[b]{1\textwidth}
        \centering
        \includegraphics[width=0.7\textwidth]{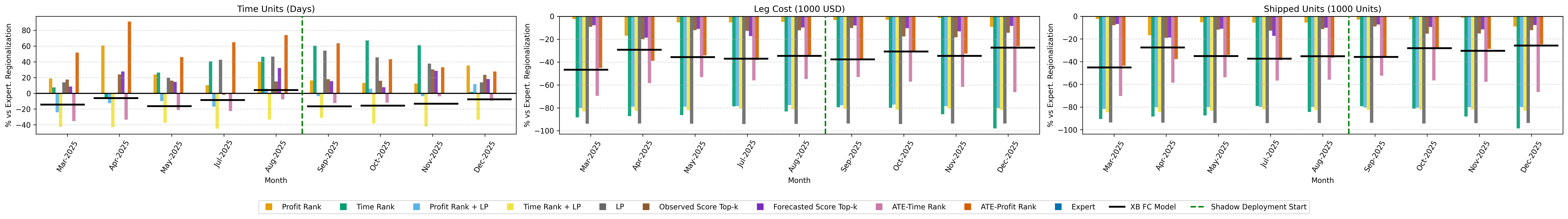}
        \caption{US$>$AU}
        \label{fig:graph3}
    \end{subfigure}
    
    \caption{Aggregate Performance - Expert vs. FC Selection Strategies: Percentage Difference in Operational Metrics}
    \label{fig:three_graphs}
\end{figure*}

\begin{figure*}[t]
    \centering
    \captionsetup{font=footnotesize}

    \begin{minipage}[b]{0.32\textwidth}
        \centering
        \includegraphics[width=\linewidth]{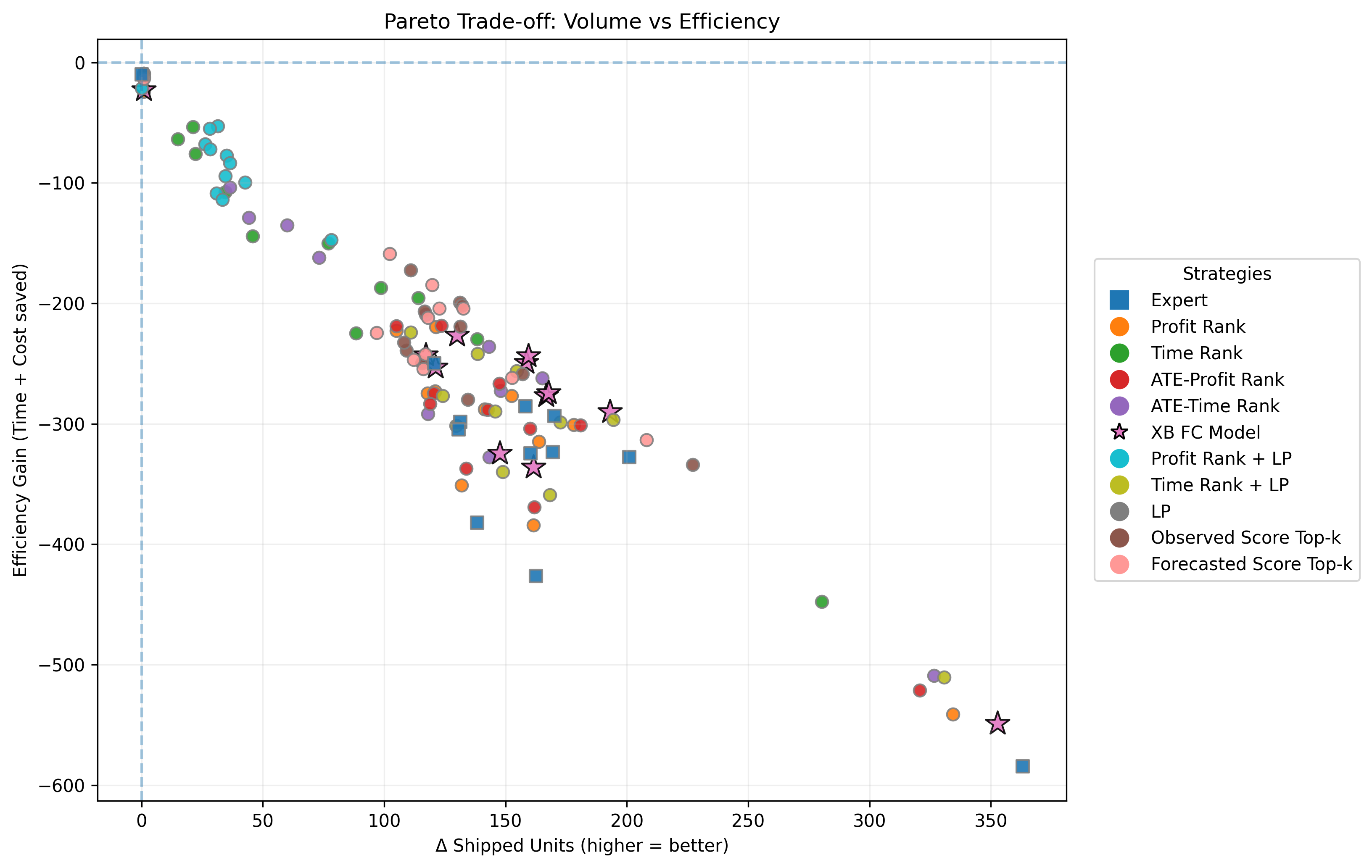}
        \small (a) US$>$SG
    \end{minipage}
    \hfill
    \begin{minipage}[b]{0.32\textwidth}
        \centering
        \includegraphics[width=\linewidth]{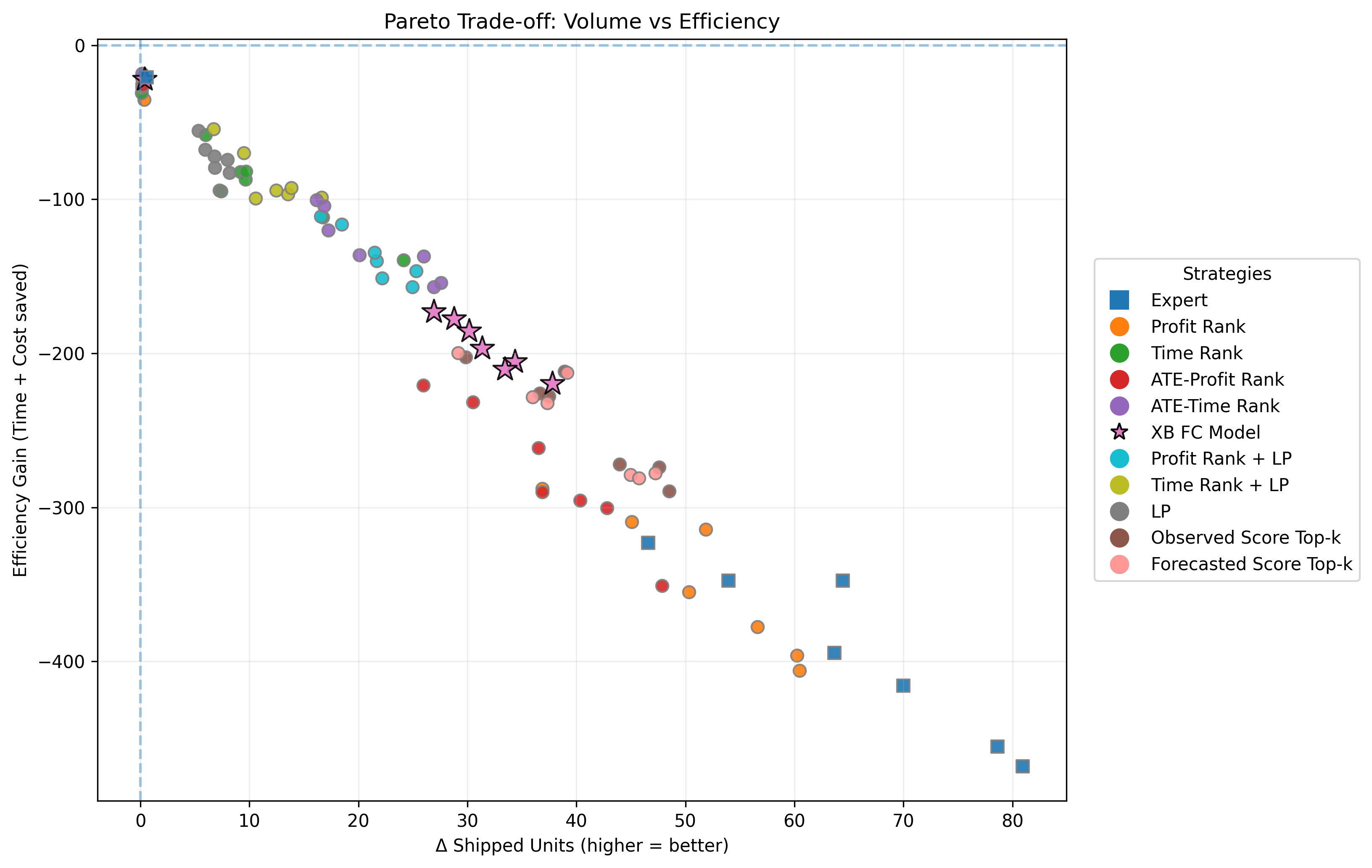}
        
        \small(b) US$>$DE
    \end{minipage}
    \hfill
    \begin{minipage}[b]{0.32\textwidth}
        \centering
        \includegraphics[width=\linewidth]{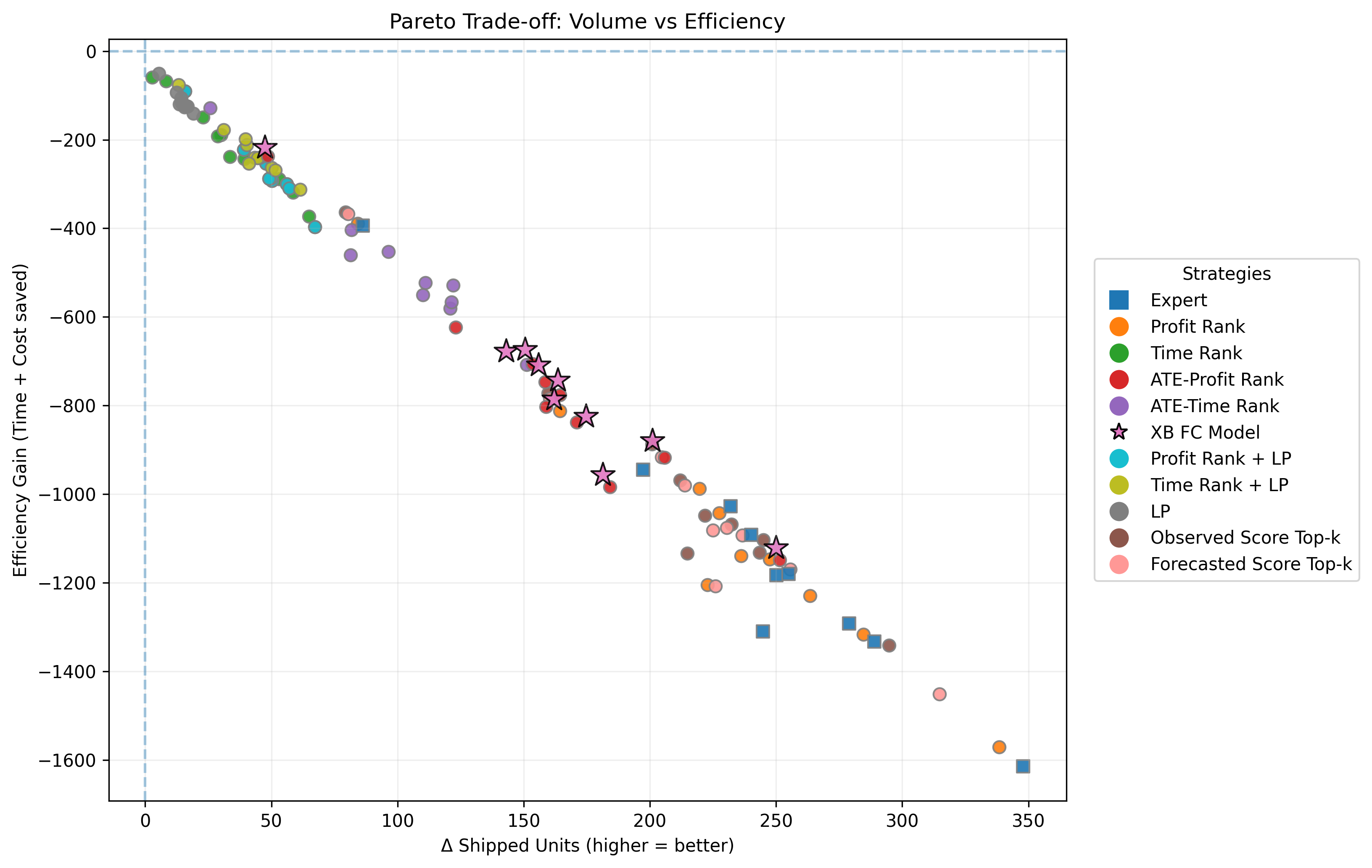}
        
        \small (c) US$>$AU
    \end{minipage}

    \vspace{2pt}

    \caption{Pareto Trade-off Between Throughput and Efficiency Across Fulfillment Strategies (Monthly Absolute Values). Each point corresponds to a strategy for a month, visualizing the Pareto frontier between shipment volume and operational efficiency across FC selection strategies versus the no-regionalization baseline.}
    \label{fig:three_graphs}
\end{figure*}

\begin{figure*}[t]
    \centering
    \captionsetup{font=footnotesize}

    \parbox[b]{0.32\textwidth}{
        \centering
        \includegraphics[width=\linewidth]{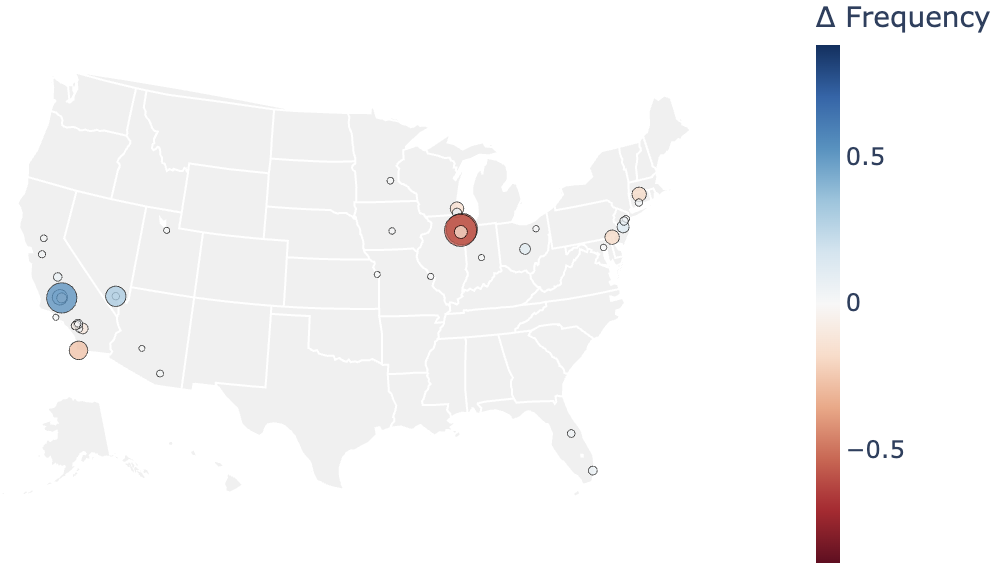}\\
        \small (a) US$>$SG
    }
    \hfill
    \parbox[b]{0.32\textwidth}{
        \centering
        \includegraphics[width=\linewidth]{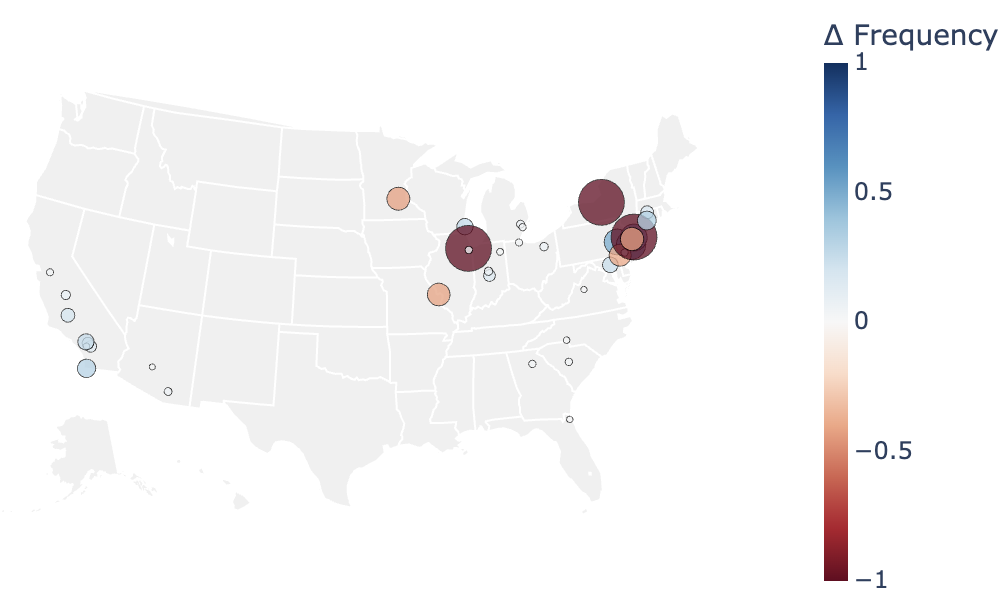}\\
        \small (b) US$>$DE
    }
    \hfill
    \parbox[b]{0.32\textwidth}{
        \centering
        \includegraphics[width=\linewidth]{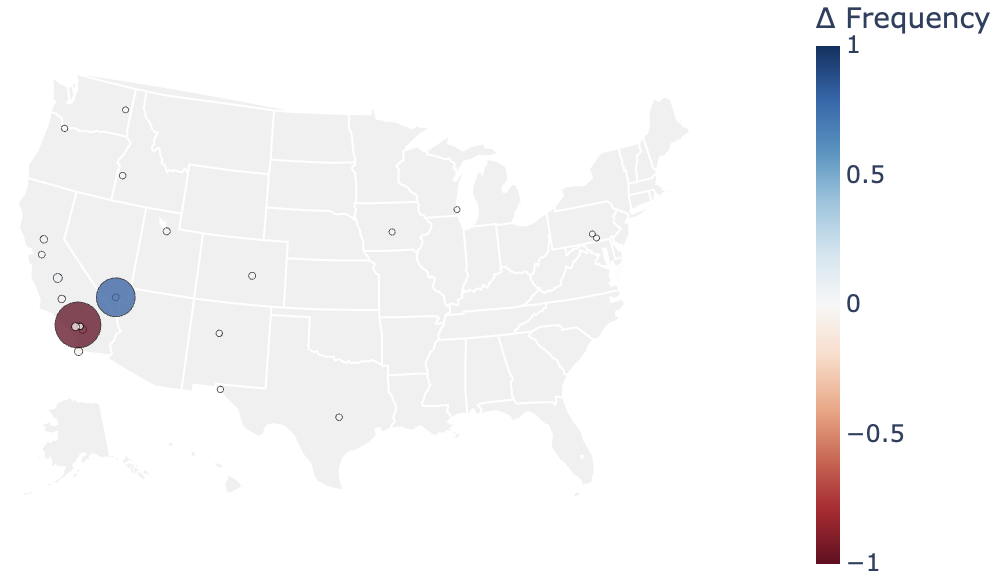}\\
        \small (c) US$>$AU
    }

    \caption{Spatial comparison of XB FC model selection frequency relative to the Expert baseline. Each point represents a warehouse location; marker size is proportional to deviation magnitude; red indicates higher-than-expert selection, while blue indicates lower-than-expert selection.}
    \label{fig:three_graphs}
\end{figure*}

\textbf{4.1 Aggregate Performance:} \textbf{4.1.1 Against No Regionalization Baseline.} All strategies achieve >95\% time reduction, confirming the fundamental value of FC selection (Appendix Table 3). Strategies separate sharply on cost versus volume: LP maximizes cost reductions (88–91\%) but at near-total volume loss (87–90\%), while the Expert policy achieves the lowest volume reduction (36–61\%) with modest cost gains (46–63\%). The XB FC Model balances these extremes: 51–80\% cost reduction at 41–79\% volume reduction, consistently improving upon Expert in cost efficiency without proportional volume loss.

\textbf{4.1.2 Against Expert Baseline:}
The trend observed in Section 4.1.1 continues here. Relative to Expert selection, the XB FC Selection Model provides the best cross-corridor balance between efficiency and volume retention (Table 1; Figure 3). The main baselines exhibit sharper trade-offs: Single-objective rankings show the sharpest trade-offs---Profit Rank preserves volume but increases delivery time in every corridor (up to $+40.5\%$), while Time Rank achieves large time reductions for US$>$SG and US$>$DE but increases delivery time for US$>$AU and incurs severe volume loss (46--87\%). ATE-based candidate generation yields marginal improvements over single-objective ranking counterparts but does not resolve these trade-offs. All three LP variants produce substantial time reductions for US$>$SG and US$>$DE but diverge sharply on volume retention, with Vanilla LP incurring 83--94\% volume loss for US$>$DE and US$>$AU. The composite-score heuristics occupy a moderate band but increase delivery time for US$>$AU by 14.8--21.9\%, demonstrating that static scoring rules cannot capture corridor-specific structure. By contrast, the XB FC Model is one of only three strategies that achieve delivery-time reductions across all three corridors---and it does so at substantially lower volume cost than either alternative. Specifically, the model achieves 24.4--41.8\% / 27.1--51.1\% / 8.5--13.3\% time reductions and 4.4--8.6\% / 45.9--48.3\% / 32.6--36.4\% cost reductions for US$>$SG / US$>$DE / US$>$AU, respectively, while limiting shipped-unit reallocation to $-7.7\%$ to $+2.5\%$ / 48.5--48.9\% / 30.1--35.9\%. These results are stable across pre- and post-shadow-deployment windows and statistically significant (Wilcoxon $p < 0.02$; paired $t$-test $p < 0.005$; Cohen's $d$: $-1.03$ to $-3.89$; see Appendix Tables 4--9).\textbf{Geographic Distribution of Selected FCs:} Expert and XB FC Model selections broadly overlap in coastal orientation for each arc (see Appendix Figure 6), but Figure 5 reveals where they diverge.

\textbf{4.1.3 Pareto Trade-Off Analysis} (see Figure 4). To characterize multi-objective efficiency, we visualize the trade-off between operational efficiency gains and shipped volume retention relative to the no-regionalization baseline via Pareto frontier plots. Strategies separate into four structural clusters: (1) \textit{High-efficiency, high-volume-loss} (upper-left): Time Rank, LP variants, and ATE-Time Rank achieve the largest time and cost reductions at steep volume penalties. (2) \textit{Low-efficiency, low-volume-loss} (lower-right): Profit Rank and ATE-Profit Rank preserve volume but offer minimal or negative time improvement. (3) \textit{Moderate} (center): Composite-score heuristics achieve intermediate trade-offs for US$>$SG and US$>$DE but drift into the low-efficiency cluster for US$>$AU. (4) \textit{Pareto-optimal} (frontier): The XB FC Model consistently occupies the Pareto frontier — the region of maximum efficiency gain per unit of volume reduction — across all three corridors. No other strategy achieves this cross-corridor consistency.

\textbf{4.2 Counterfactual Evaluation: Shipment-Level Performance.} Shipment-level counterfactual matching corroborates the aggregate results (Tables 2). Under both same product and nearest neighbor settings, the XB FC Model achieves the lowest or near-lowest mean delivery time and shipping cost across all three corridors, and the highest 5/5 target compliance for US$>$SG and US$>$DE. For US$>$AU, the results diverge: Expert leads under nearest-neighbor matching (0.93 vs. 1.13 days), but the same-ASIN estimator — which controls for product-level heterogeneity — favors XB FC on every metric (1.84 vs. 2.31 days, 8.21\% vs. 6.47\% 5/5 compliance). Across both matching methods, the XB FC Model consistently outperforms LP-based selection on all metrics and corridors, demonstrating the incremental value of DRE-based candidate generation alongside the multi-objective LP. Paired t-tests on matched samples (Appendix Tables 10–12) confirm all comparative differences are statistically significant at the 1\% level.

\section{Conclusion and Limitations}

We presented the XB FC Selection Model, which combines forecasting, doubly robust candidate generation, and multi-objective LP for FC selection under soft regionalization. Across three production corridors, the model consistently provides a stronger efficiency–volume trade-off than expert, ranking, and LP baselines, reducing delivery time by 8.5–51.1\% and cost by up to 48.3\% relative to Expert while improving 5/5 compliance by up to 6.5 percentage points. \textbf{Limitations}: The main limitations are the lack of direct FC-level XB inventory/capacity observations and the inability to conduct online A/B tests because only one policy can be active per corridor. Furthermore, optimal FC selection requires substantial inventory reallocation (ex: ~48\% for US$>$DE), highlighting a gap with current placement. To address these limitations, future work will incorporate explicit inventory data, jointly optimize FC selection and physical inventory reallocation via reinforcement learning, and report post-deployment results.

\clearpage
\FloatBarrier
\appendix
\section*{Appendix}




\begin{table*}[t]
\centering
\scriptsize
\setlength{\tabcolsep}{3pt}

\caption{FC Selection Strategies Comparison Across Regions -- Historical Performance Relative to No Regionalization (\% Values).}
\label{tab:fc_comparison_no_reg_all_regions}

\begin{tabular}{l|rrr|rrr|rrr|rrr|rrr|rrr}
\toprule

& \multicolumn{6}{c|}{US$>$SG}
& \multicolumn{6}{c|}{US$>$DE}
& \multicolumn{6}{c}{US$>$AU} \\

& \multicolumn{3}{c|}{Pre-Deployment}
& \multicolumn{3}{c|}{Post-Deployment}
& \multicolumn{3}{c|}{Pre-Deployment}
& \multicolumn{3}{c|}{Post-Deployment}
& \multicolumn{3}{c|}{Pre-Deployment}
& \multicolumn{3}{c}{Post-Deployment} \\

\cmidrule(lr){2-4}
\cmidrule(lr){5-7}
\cmidrule(lr){8-10}
\cmidrule(lr){11-13}
\cmidrule(lr){14-16}
\cmidrule(lr){17-19}

Strategy
& Time & Cost & Units
& Time & Cost & Units
& Time & Cost & Units
& Time & Cost & Units
& Time & Cost & Units
& Time & Cost & Units \\

\midrule

Expert
& -0.959 & -0.452 & -0.331
& -0.983 & -0.462 & -0.328
& -0.955 & -0.641 & -0.582
& -0.957 & -0.628 & -0.609
& -0.991 & -0.519 & -0.441
& -0.992 & -0.456 & -0.378 \\

Profit Rank
& -0.957 & -0.507 & -0.350
& -0.981 & -0.531 & -0.408
& -0.947 & -0.713 & -0.670
& -0.953 & -0.703 & -0.693
& -0.988 & -0.553 & -0.480
& -0.991 & -0.479 & -0.402 \\

Time Rank
& -0.980 & -0.731 & -0.672
& -0.986 & -0.765 & -0.702
& -0.980 & -0.924 & -0.912
& -0.986 & -0.941 & -0.945
& -0.988 & -0.925 & -0.919
& -0.989 & -0.919 & -0.912 \\

ATE-Profit Rank
& -0.962 & -0.514 & -0.352
& -0.982 & -0.532 & -0.411
& -0.949 & -0.788 & -0.755
& -0.958 & -0.794 & -0.786
& -0.985 & -0.695 & -0.645
& -0.989 & -0.627 & -0.560 \\

ATE-Time Rank
& -0.976 & -0.540 & -0.418
& -0.988 & -0.659 & -0.573
& -0.980 & -0.890 & -0.861
& -0.986 & -0.884 & -0.871
& -0.993 & -0.792 & -0.762
& -0.993 & -0.778 & -0.738 \\

LP
& -0.978 & -0.448 & -0.269
& -0.987 & -0.490 & -0.309
& -0.980 & -0.958 & -0.956
& -0.984 & -0.957 & -0.958
& -0.988 & -0.971 & -0.967
& -0.989 & -0.967 & -0.964 \\

XB FC Model (ATE-LP)
& -0.983 & -0.492 & -0.322
& -0.989 & -0.520 & -0.345
& -0.976 & -0.829 & -0.808
& -0.980 & -0.804 & -0.794
& -0.992 & -0.689 & -0.637
& -0.993 & -0.632 & -0.565 \\

Profit Rank + LP
& -0.981 & -0.869 & -0.859
& -0.986 & -0.903 & -0.857
& -0.980 & -0.882 & -0.863
& -0.985 & -0.876 & -0.870
& -0.992 & -0.898 & -0.889
& -0.992 & -0.882 & -0.878 \\

Time Rank + LP
& -0.977 & -0.476 & -0.303
& -0.988 & -0.507 & -0.377
& -0.983 & -0.927 & -0.917
& -0.989 & -0.937 & -0.938
& -0.995 & -0.912 & -0.905
& -0.995 & -0.898 & -0.891 \\

Observed Score Top-k
& -0.984 & -0.608 & -0.467
& -0.988 & -0.624 & -0.472
& -0.970 & -0.787 & -0.747
& -0.972 & -0.767 & -0.751
& -0.990 & -0.586 & -0.514
& -0.991 & -0.539 & -0.460 \\

Forecasted Score Top-k
& -0.985 & -0.646 & -0.522
& -0.988 & -0.628 & -0.486
& -0.969 & -0.784 & -0.746
& -0.972 & -0.772 & -0.760
& -0.989 & -0.584 & -0.515
& -0.991 & -0.510 & -0.434 \\

\bottomrule
\end{tabular}
\end{table*}

\begin{table}
\caption{US$>$SG: Statistical Comparison of FC Selection Strategies vs. Expert Baseline}
\centering
\resizebox{\linewidth}{!}{
\begin{tabular}{lllrrrr}
\toprule
 & Baseline & Strategy & Mean Difference & Wilcoxon p-value & T-test p-value & Cohen's d \\
Metric &  &  &  &  &  &  \\
\midrule
Time Units (Days) & Expert & ATE-Profit Rank & 0.717 & 0.569 & 0.888 & 0.042 \\
Leg Cost (1000 USD) & Expert & ATE-Profit Rank & -66.354 & 0.001 & 0.000 & -1.775 \\
Shipped Units (1000 Units) & Expert & ATE-Profit Rank & -9.257 & 0.052 & 0.053 & -0.626 \\
Time Units (Days) & Expert & ATE-Time Rank & -26.363 & 0.001 & 0.005 & -1.021 \\
Leg Cost (1000 USD) & Expert & ATE-Time Rank & -162.802 & 0.001 & 0.000 & -1.426 \\
Shipped Units (1000 Units) & Expert & ATE-Time Rank & -36.753 & 0.001 & 0.001 & -1.391 \\
Time Units (Days) & Expert & Forecasted Score Top-k & -33.457 & 0.042 & 0.017 & -0.814 \\
Leg Cost (1000 USD) & Expert & Forecasted Score Top-k & -181.416 & 0.001 & 0.000 & -1.454 \\
Shipped Units (1000 Units) & Expert & Forecasted Score Top-k & -42.215 & 0.001 & 0.007 & -0.963 \\
Time Units (Days) & Expert & Observed Score Top-k & -34.390 & 0.003 & 0.012 & -0.868 \\
Leg Cost (1000 USD) & Expert & Observed Score Top-k & -166.891 & 0.001 & 0.000 & -1.549 \\
Shipped Units (1000 Units) & Expert & Observed Score Top-k & -35.885 & 0.001 & 0.008 & -0.943 \\
Time Units (Days) & Expert & Profit Rank & 9.849 & 0.027 & 0.021 & 0.777 \\
Leg Cost (1000 USD) & Expert & Profit Rank & -63.126 & 0.001 & 0.000 & -2.100 \\
Shipped Units (1000 Units) & Expert & Profit Rank & -8.334 & 0.034 & 0.034 & -0.699 \\
Time Units (Days) & Expert & Profit Rank + LP & -25.284 & 0.021 & 0.034 & -0.697 \\
Leg Cost (1000 USD) & Expert & Profit Rank + LP & -444.192 & 0.001 & 0.000 & -2.351 \\
Shipped Units (1000 Units) & Expert & Profit Rank + LP & -125.011 & 0.001 & 0.000 & -1.959 \\
Time Units (Days) & Expert & Time Rank & -22.943 & 0.042 & 0.044 & -0.655 \\
Leg Cost (1000 USD) & Expert & Time Rank & -293.775 & 0.001 & 0.000 & -2.078 \\
Shipped Units (1000 Units) & Expert & Time Rank & -77.631 & 0.001 & 0.000 & -2.447 \\
Time Units (Days) & Expert & Time Rank + LP & -34.626 & 0.007 & 0.007 & -0.946 \\
Leg Cost (1000 USD) & Expert & Time Rank + LP & -52.239 & 0.001 & 0.000 & -1.707 \\
Shipped Units (1000 Units) & Expert & Time Rank + LP & -2.404 & 0.266 & 0.338 & -0.289 \\
Time Units (Days) & Expert & LP & -24.414 & 0.007 & 0.021 & -0.774 \\
Leg Cost (1000 USD) & Expert & LP & -18.058 & 0.012 & 0.013 & -0.850 \\
Shipped Units (1000 Units) & Expert & LP & 6.539 & 0.064 & 0.041 & 0.668 \\
Time Units (Days) & Expert & XB FC Model & -25.758 & 0.001 & 0.006 & -0.969 \\
Leg Cost (1000 USD) & Expert & XB FC Model & -37.739 & 0.001 & 0.002 & -1.149 \\
Shipped Units (1000 Units) & Expert & XB FC Model & 0.321 & 0.622 & 0.939 & 0.023 \\
\bottomrule
\end{tabular}
}
\end{table}

\begin{table}
\caption{US$>$DE: Statistical Comparison of FC Selection Strategies vs. Expert Baseline}
\centering
\resizebox{\linewidth}{!}{
\begin{tabular}{lllrrrr}
\toprule
 & Baseline & Strategy & Mean Difference & Wilcoxon p-value & T-test p-value & Cohen's d \\
Metric &  &  &  &  &  &  \\
\midrule
Time Units (Days) & Expert & ATE-Profit Rank & 14.461 & 0.109 & 0.092 & 0.690 \\
Leg Cost (1000 USD) & Expert & ATE-Profit Rank & -213.611 & 0.008 & 0.000 & -2.286 \\
Shipped Units (1000 Units) & Expert & ATE-Profit Rank & -24.708 & 0.008 & 0.000 & -2.219 \\
Time Units (Days) & Expert & ATE-Time Rank & -113.836 & 0.008 & 0.000 & -2.245 \\
Leg Cost (1000 USD) & Expert & ATE-Time Rank & -347.407 & 0.008 & 0.000 & -2.303 \\
Shipped Units (1000 Units) & Expert & ATE-Time Rank & -38.476 & 0.008 & 0.000 & -2.218 \\
Time Units (Days) & Expert & Forecasted Score Top-k & -62.847 & 0.016 & 0.001 & -2.010 \\
Leg Cost (1000 USD) & Expert & Forecasted Score Top-k & -197.450 & 0.008 & 0.000 & -2.204 \\
Shipped Units (1000 Units) & Expert & Forecasted Score Top-k & -22.380 & 0.008 & 0.001 & -2.090 \\
Time Units (Days) & Expert & Observed Score Top-k & -64.021 & 0.016 & 0.001 & -1.930 \\
Leg Cost (1000 USD) & Expert & Observed Score Top-k & -197.967 & 0.008 & 0.000 & -2.257 \\
Shipped Units (1000 Units) & Expert & Observed Score Top-k & -21.960 & 0.008 & 0.001 & -2.115 \\
Time Units (Days) & Expert & Profit Rank & 27.974 & 0.008 & 0.000 & 2.443 \\
Leg Cost (1000 USD) & Expert & Profit Rank & -100.810 & 0.008 & 0.001 & -2.053 \\
Shipped Units (1000 Units) & Expert & Profit Rank & -12.122 & 0.008 & 0.001 & -1.948 \\
Time Units (Days) & Expert & Profit Rank + LP & -111.600 & 0.016 & 0.001 & -1.924 \\
Leg Cost (1000 USD) & Expert & Profit Rank + LP & -336.185 & 0.008 & 0.000 & -2.245 \\
Shipped Units (1000 Units) & Expert & Profit Rank + LP & -38.491 & 0.008 & 0.000 & -2.179 \\
Time Units (Days) & Expert & Time Rank & -114.044 & 0.016 & 0.001 & -1.817 \\
Leg Cost (1000 USD) & Expert & Time Rank & -407.670 & 0.008 & 0.000 & -2.184 \\
Shipped Units (1000 Units) & Expert & Time Rank & -46.993 & 0.008 & 0.001 & -2.141 \\
Time Units (Days) & Expert & Time Rank + LP & -128.606 & 0.016 & 0.001 & -1.962 \\
Leg Cost (1000 USD) & Expert & Time Rank + LP & -407.038 & 0.008 & 0.000 & -2.295 \\
Shipped Units (1000 Units) & Expert & Time Rank + LP & -46.922 & 0.008 & 0.000 & -2.233 \\
Time Units (Days) & Expert & LP & -111.850 & 0.016 & 0.001 & -1.820 \\
Leg Cost (1000 USD) & Expert & LP & -443.032 & 0.008 & 0.000 & -2.298 \\
Shipped Units (1000 Units) & Expert & LP & -51.293 & 0.008 & 0.000 & -2.219 \\
Time Units (Days) & Expert & XB FC Model & -93.582 & 0.016 & 0.001 & -2.052 \\
Leg Cost (1000 USD) & Expert & XB FC Model & -251.973 & 0.008 & 0.001 & -2.090 \\
Shipped Units (1000 Units) & Expert & XB FC Model & -29.458 & 0.008 & 0.001 & -2.003 \\
\bottomrule
\end{tabular}
}
\end{table}

\begin{table}
\caption{US$>$AU: Statistical Comparison of FC Selection Strategies vs. Expert Baseline}
\centering
\resizebox{\linewidth}{!}{
\begin{tabular}{lllrrrr}
\toprule
 & Baseline & Strategy & Mean Difference & Wilcoxon p-value & T-test p-value & Cohen's d \\
Metric &  &  &  &  &  &  \\
\midrule
Time Units (Days) & Expert & ATE-Profit Rank & 0.000 & 0.002 & 0.000 & 2.359 \\
Leg Cost (1000 USD) & Expert & ATE-Profit Rank & -0.753 & 0.002 & 0.000 & -4.143 \\
Shipped Units (1000 Units) & Expert & ATE-Profit Rank & -0.080 & 0.002 & 0.000 & -4.061 \\
Time Units (Days) & Expert & ATE-Time Rank & -0.000 & 0.006 & 0.005 & -1.170 \\
Leg Cost (1000 USD) & Expert & ATE-Time Rank & -1.284 & 0.002 & 0.000 & -3.453 \\
Shipped Units (1000 Units) & Expert & ATE-Time Rank & -0.140 & 0.002 & 0.000 & -3.847 \\
Time Units (Days) & Expert & Forecasted Score Top-k & 0.000 & 0.004 & 0.001 & 1.607 \\
Leg Cost (1000 USD) & Expert & Forecasted Score Top-k & -0.261 & 0.002 & 0.000 & -2.476 \\
Shipped Units (1000 Units) & Expert & Forecasted Score Top-k & -0.027 & 0.002 & 0.000 & -2.342 \\
Time Units (Days) & Expert & Observed Score Top-k & 0.000 & 0.004 & 0.000 & 2.007 \\
Leg Cost (1000 USD) & Expert & Observed Score Top-k & -0.322 & 0.002 & 0.000 & -2.399 \\
Shipped Units (1000 Units) & Expert & Observed Score Top-k & -0.032 & 0.002 & 0.000 & -2.563 \\
Time Units (Days) & Expert & Profit Rank & 0.000 & 0.002 & 0.001 & 1.524 \\
Leg Cost (1000 USD) & Expert & Profit Rank & -0.124 & 0.002 & 0.002 & -1.417 \\
Shipped Units (1000 Units) & Expert & Profit Rank & -0.013 & 0.002 & 0.001 & -1.481 \\
Time Units (Days) & Expert & Profit Rank + LP & -0.000 & 0.695 & 0.610 & -0.167 \\
Leg Cost (1000 USD) & Expert & Profit Rank + LP & -1.736 & 0.002 & 0.000 & -3.542 \\
Shipped Units (1000 Units) & Expert & Profit Rank + LP & -0.194 & 0.002 & 0.000 & -3.580 \\
Time Units (Days) & Expert & Time Rank & 0.000 & 0.010 & 0.004 & 1.224 \\
Leg Cost (1000 USD) & Expert & Time Rank & -1.876 & 0.002 & 0.000 & -3.528 \\
Shipped Units (1000 Units) & Expert & Time Rank & -0.208 & 0.002 & 0.000 & -3.777 \\
Time Units (Days) & Expert & Time Rank + LP & -0.000 & 0.002 & 0.000 & -4.845 \\
Leg Cost (1000 USD) & Expert & Time Rank + LP & -1.801 & 0.002 & 0.000 & -3.522 \\
Shipped Units (1000 Units) & Expert & Time Rank + LP & -0.200 & 0.002 & 0.000 & -3.646 \\
Time Units (Days) & Expert & LP & 0.000 & 0.002 & 0.002 & 1.321 \\
Leg Cost (1000 USD) & Expert & LP & -2.074 & 0.002 & 0.000 & -3.466 \\
Shipped Units (1000 Units) & Expert & LP & -0.228 & 0.002 & 0.000 & -3.548 \\
Time Units (Days) & Expert & XB FC Model & -0.000 & 0.004 & 0.001 & -1.596 \\
Leg Cost (1000 USD) & Expert & XB FC Model & -0.750 & 0.002 & 0.000 & -3.888 \\
Shipped Units (1000 Units) & Expert & XB FC Model & -0.079 & 0.002 & 0.000 & -3.847 \\
\bottomrule
\end{tabular}
}
\end{table}

\begin{table}
\caption{US$>$SG: Statistical Comparison of Algorithmic FC Selection Strategies vs. XB FC Selection Model}
\centering
\resizebox{\linewidth}{!}{
\begin{tabular}{lllrrrr}
\toprule
 & Baseline & Strategy & Mean Difference & Wilcoxon p-value & T-test p-value & Cohen's d \\
Metric &  &  &  &  &  &  \\
\midrule
Time Units (Days) & XB FC Model & ATE-Profit Rank & 26.475 & 0.000 & 0.001 & 1.216 \\
Leg Cost (1000 USD) & XB FC Model & ATE-Profit Rank & -28.615 & 0.001 & 0.000 & -1.705 \\
Shipped Units (1000 Units) & XB FC Model & ATE-Profit Rank & -9.578 & 0.000 & 0.000 & -2.104 \\
Time Units (Days) & XB FC Model & ATE-Time Rank & -0.606 & 0.910 & 0.710 & -0.110 \\
Leg Cost (1000 USD) & XB FC Model & ATE-Time Rank & -125.063 & 0.000 & 0.004 & -1.033 \\
Shipped Units (1000 Units) & XB FC Model & ATE-Time Rank & -37.074 & 0.000 & 0.001 & -1.226 \\
Time Units (Days) & XB FC Model & Forecasted Score Top-k & -7.699 & 0.176 & 0.146 & -0.452 \\
Leg Cost (1000 USD) & XB FC Model & Forecasted Score Top-k & -143.677 & 0.000 & 0.000 & -1.441 \\
Shipped Units (1000 Units) & XB FC Model & Forecasted Score Top-k & -42.536 & 0.000 & 0.001 & -1.286 \\
Time Units (Days) & XB FC Model & Observed Score Top-k & -8.633 & 0.129 & 0.087 & -0.543 \\
Leg Cost (1000 USD) & XB FC Model & Observed Score Top-k & -129.152 & 0.000 & 0.000 & -1.596 \\
Shipped Units (1000 Units) & XB FC Model & Observed Score Top-k & -36.207 & 0.000 & 0.001 & -1.355 \\
Time Units (Days) & XB FC Model & Profit Rank & 35.607 & 0.000 & 0.000 & 1.431 \\
Leg Cost (1000 USD) & XB FC Model & Profit Rank & -25.387 & 0.012 & 0.006 & -0.974 \\
Shipped Units (1000 Units) & XB FC Model & Profit Rank & -8.655 & 0.002 & 0.001 & -1.260 \\
Time Units (Days) & XB FC Model & Profit Rank + LP & 0.474 & 0.970 & 0.917 & 0.031 \\
Leg Cost (1000 USD) & XB FC Model & Profit Rank + LP & -406.453 & 0.000 & 0.000 & -2.480 \\
Shipped Units (1000 Units) & XB FC Model & Profit Rank + LP & -125.332 & 0.000 & 0.000 & -2.235 \\
Time Units (Days) & XB FC Model & Time Rank & 2.814 & 0.519 & 0.494 & 0.204 \\
Leg Cost (1000 USD) & XB FC Model & Time Rank & -256.035 & 0.001 & 0.000 & -1.758 \\
Shipped Units (1000 Units) & XB FC Model & Time Rank & -77.952 & 0.000 & 0.000 & -2.047 \\
Time Units (Days) & XB FC Model & Time Rank + LP & -8.869 & 0.064 & 0.060 & -0.604 \\
Leg Cost (1000 USD) & XB FC Model & Time Rank + LP & -14.500 & 0.110 & 0.272 & -0.333 \\
Shipped Units (1000 Units) & XB FC Model & Time Rank + LP & -2.725 & 0.129 & 0.356 & -0.278 \\
Time Units (Days) & XB FC Model & LP & 1.343 & 0.970 & 0.630 & 0.143 \\
Leg Cost (1000 USD) & XB FC Model & LP & 19.681 & 0.007 & 0.038 & 0.680 \\
Shipped Units (1000 Units) & XB FC Model & LP & 6.218 & 0.000 & 0.012 & 0.862 \\
\bottomrule
\end{tabular}
}
\end{table}

\begin{table}
\caption{US$>$DE: Statistical Comparison of Algorithmic FC Selection Strategies vs. XB FC Selection Model}
\centering
\resizebox{\linewidth}{!}{
\begin{tabular}{lllrrrr}
\toprule
 & Baseline & Strategy & Mean Difference & Wilcoxon p-value & T-test p-value & Cohen's d \\
Metric &  &  &  &  &  &  \\
\midrule
Time Units (Days) & XB FC Model & ATE-Profit Rank & 108.042 & 0.008 & 0.000 & 2.182 \\
Leg Cost (1000 USD) & XB FC Model & ATE-Profit Rank & 38.362 & 0.109 & 0.035 & 0.923 \\
Shipped Units (1000 Units) & XB FC Model & ATE-Profit Rank & 4.750 & 0.055 & 0.034 & 0.933 \\
Time Units (Days) & XB FC Model & ATE-Time Rank & -20.254 & 0.008 & 0.000 & -2.397 \\
Leg Cost (1000 USD) & XB FC Model & ATE-Time Rank & -95.434 & 0.008 & 0.001 & -2.029 \\
Shipped Units (1000 Units) & XB FC Model & ATE-Time Rank & -9.019 & 0.008 & 0.002 & -1.675 \\
Time Units (Days) & XB FC Model & Forecasted Score Top-k & 30.735 & 0.008 & 0.002 & 1.709 \\
Leg Cost (1000 USD) & XB FC Model & Forecasted Score Top-k & 54.523 & 0.016 & 0.007 & 1.326 \\
Shipped Units (1000 Units) & XB FC Model & Forecasted Score Top-k & 7.077 & 0.016 & 0.005 & 1.409 \\
Time Units (Days) & XB FC Model & Observed Score Top-k & 29.560 & 0.008 & 0.001 & 1.819 \\
Leg Cost (1000 USD) & XB FC Model & Observed Score Top-k & 54.006 & 0.016 & 0.007 & 1.328 \\
Shipped Units (1000 Units) & XB FC Model & Observed Score Top-k & 7.497 & 0.016 & 0.003 & 1.527 \\
Time Units (Days) & XB FC Model & Profit Rank & 121.555 & 0.008 & 0.000 & 2.554 \\
Leg Cost (1000 USD) & XB FC Model & Profit Rank & 151.163 & 0.008 & 0.001 & 2.032 \\
Shipped Units (1000 Units) & XB FC Model & Profit Rank & 17.335 & 0.016 & 0.001 & 1.937 \\
Time Units (Days) & XB FC Model & Profit Rank + LP & -18.018 & 0.023 & 0.011 & -1.222 \\
Leg Cost (1000 USD) & XB FC Model & Profit Rank + LP & -84.212 & 0.008 & 0.000 & -2.236 \\
Shipped Units (1000 Units) & XB FC Model & Profit Rank + LP & -9.034 & 0.008 & 0.000 & -2.225 \\
Time Units (Days) & XB FC Model & Time Rank & -20.462 & 0.023 & 0.021 & -1.045 \\
Leg Cost (1000 USD) & XB FC Model & Time Rank & -155.697 & 0.008 & 0.001 & -1.890 \\
Shipped Units (1000 Units) & XB FC Model & Time Rank & -17.535 & 0.008 & 0.002 & -1.772 \\
Time Units (Days) & XB FC Model & Time Rank + LP & -35.024 & 0.016 & 0.002 & -1.644 \\
Leg Cost (1000 USD) & XB FC Model & Time Rank + LP & -155.065 & 0.008 & 0.000 & -2.347 \\
Shipped Units (1000 Units) & XB FC Model & Time Rank + LP & -17.465 & 0.008 & 0.000 & -2.319 \\
Time Units (Days) & XB FC Model & LP & -18.269 & 0.023 & 0.026 & -0.993 \\
Leg Cost (1000 USD) & XB FC Model & LP & -191.059 & 0.008 & 0.000 & -2.422 \\
Shipped Units (1000 Units) & XB FC Model & LP & -21.836 & 0.008 & 0.000 & -2.391 \\
\bottomrule
\end{tabular}
}
\end{table}

\begin{table}
\caption{US$>$AU: Statistical Comparison of Algorithmic FC Selection Strategies vs. XB FC Selection Model}
\centering
\resizebox{\linewidth}{!}{
\begin{tabular}{lllrrrr}
\toprule
 & Baseline & Strategy & Mean Difference & Wilcoxon p-value & T-test p-value & Cohen's d \\
Metric &  &  &  &  &  &  \\
\midrule
Time Units (Days) & XB FC Model & ATE-Profit Rank & 0.000 & 0.002 & 0.000 & 2.975 \\
Leg Cost (1000 USD) & XB FC Model & ATE-Profit Rank & -0.003 & 0.375 & 0.873 & -0.052 \\
Shipped Units (1000 Units) & XB FC Model & ATE-Profit Rank & -0.001 & 0.846 & 0.678 & -0.136 \\
Time Units (Days) & XB FC Model & ATE-Time Rank & -0.000 & 0.432 & 0.328 & -0.327 \\
Leg Cost (1000 USD) & XB FC Model & ATE-Time Rank & -0.534 & 0.002 & 0.000 & -2.197 \\
Shipped Units (1000 Units) & XB FC Model & ATE-Time Rank & -0.061 & 0.002 & 0.000 & -2.429 \\
Time Units (Days) & XB FC Model & Forecasted Score Top-k & 0.000 & 0.002 & 0.000 & 2.993 \\
Leg Cost (1000 USD) & XB FC Model & Forecasted Score Top-k & 0.489 & 0.002 & 0.000 & 3.272 \\
Shipped Units (1000 Units) & XB FC Model & Forecasted Score Top-k & 0.052 & 0.002 & 0.000 & 3.169 \\
Time Units (Days) & XB FC Model & Observed Score Top-k & 0.000 & 0.002 & 0.000 & 2.557 \\
Leg Cost (1000 USD) & XB FC Model & Observed Score Top-k & 0.428 & 0.002 & 0.000 & 2.990 \\
Shipped Units (1000 Units) & XB FC Model & Observed Score Top-k & 0.047 & 0.002 & 0.000 & 2.822 \\
Time Units (Days) & XB FC Model & Profit Rank & 0.000 & 0.002 & 0.000 & 2.495 \\
Leg Cost (1000 USD) & XB FC Model & Profit Rank & 0.626 & 0.002 & 0.000 & 2.790 \\
Shipped Units (1000 Units) & XB FC Model & Profit Rank & 0.066 & 0.002 & 0.000 & 2.738 \\
Time Units (Days) & XB FC Model & Profit Rank + LP & 0.000 & 0.105 & 0.087 & 0.607 \\
Leg Cost (1000 USD) & XB FC Model & Profit Rank + LP & -0.986 & 0.002 & 0.000 & -3.025 \\
Shipped Units (1000 Units) & XB FC Model & Profit Rank + LP & -0.115 & 0.002 & 0.000 & -3.032 \\
Time Units (Days) & XB FC Model & Time Rank & 0.000 & 0.002 & 0.001 & 1.450 \\
Leg Cost (1000 USD) & XB FC Model & Time Rank & -1.127 & 0.002 & 0.000 & -2.864 \\
Shipped Units (1000 Units) & XB FC Model & Time Rank & -0.129 & 0.002 & 0.000 & -3.067 \\
Time Units (Days) & XB FC Model & Time Rank + LP & -0.000 & 0.002 & 0.000 & -2.903 \\
Leg Cost (1000 USD) & XB FC Model & Time Rank + LP & -1.051 & 0.002 & 0.000 & -3.036 \\
Shipped Units (1000 Units) & XB FC Model & Time Rank + LP & -0.121 & 0.002 & 0.000 & -3.125 \\
Time Units (Days) & XB FC Model & LP & 0.000 & 0.002 & 0.001 & 1.621 \\
Leg Cost (1000 USD) & XB FC Model & LP & -1.324 & 0.002 & 0.000 & -3.088 \\
Shipped Units (1000 Units) & XB FC Model & LP & -0.149 & 0.002 & 0.000 & -3.125 \\
\bottomrule
\end{tabular}
}
\end{table}

\begin{table}[htbp]
\centering
\scriptsize
\begin{tabular}{llcc}
\toprule
\textbf{Metric} & \textbf{Comparison} & \textbf{t-stat} & \textbf{p-value} \\
\midrule

\multicolumn{4}{l}{\textit{Time (Days)}} \\
\midrule
& ATE vs Actual     & -72.74 & $< 1e^{-300}$ \\
& LP vs Actual      & -70.26 & $< 1e^{-300}$ \\
& Expert vs Actual  & -72.74 & $< 1e^{-300}$ \\
& ATE vs Expert     & -78.79 & $< 1e^{-300}$ \\
& LP vs Expert      & -75.21 & $< 1e^{-300}$ \\
& ATE vs LP         & -15.73 & $1.36 \times 10^{-55}$ \\

\midrule
\multicolumn{4}{l}{\textit{Cost (USD)}} \\
\midrule
& ATE vs Actual     & -14.31 & $2.40 \times 10^{-46}$ \\
& LP vs Actual      & -4.43  & $9.48 \times 10^{-6}$  \\
& Expert vs Actual  & -14.31 & $2.40 \times 10^{-46}$ \\
& ATE vs Expert     & -16.15 & $1.57 \times 10^{-58}$ \\
& LP vs Expert      & -5.16  & $2.46 \times 10^{-7}$  \\
& ATE vs LP         & -41.71 & $< 1e^{-300}$ \\

\bottomrule
\end{tabular}
\caption{US>SG: Paired t-test results comparing delivery time and cost across strategies using matched samples.}
\label{tab:ttest_results}
\end{table}

\begin{table}[htbp]
\centering
\scriptsize
\begin{tabular}{llcc}
\toprule
\textbf{Metric} & \textbf{Comparison} & \textbf{t-stat} & \textbf{p-value} \\
\midrule

\multicolumn{4}{l}{\textit{Time (Days)}} \\
\midrule
& ATE vs Actual     & -62.50 & $< 1e^{-300}$ \\
& LP vs Actual      & -53.28 & $< 1e^{-300}$ \\
& Expert vs Actual  & -62.50 & $< 1e^{-300}$ \\
& ATE vs Expert     & -83.94 & $< 1e^{-300}$ \\
& LP vs Expert      & -72.97 & $< 1e^{-300}$ \\
& ATE vs LP         & -24.64 & $3.25 \times 10^{-133}$ \\

\midrule
\multicolumn{4}{l}{\textit{Cost (USD)}} \\
\midrule
& ATE vs Actual     & -58.91 & $< 1e^{-300}$ \\
& LP vs Actual      & -35.07 & $4.97 \times 10^{-266}$ \\
& Expert vs Actual  & -58.91 & $< 1e^{-300}$ \\
& ATE vs Expert     & -45.90 & $< 1e^{-300}$ \\
& LP vs Expert      & -22.01 & $7.26 \times 10^{-107}$ \\
& ATE vs LP         & -68.76 & $< 1e^{-300}$ \\

\bottomrule
\end{tabular}
\caption{US$>$DE: Paired t-test results comparing delivery time and cost across strategies using matched samples.}
\label{tab:ttest_results_appendix}
\end{table}

\begin{table}[htbp]
\centering
\scriptsize
\begin{tabular}{llcc}
\toprule
\textbf{Metric} & \textbf{Comparison} & \textbf{t-stat} & \textbf{p-value} \\
\midrule

\multicolumn{4}{l}{\textit{Time}} \\
\midrule
& ATE vs Actual     & -514.73 & $< 1e^{-300}$ \\
& LP vs Actual      & -507.15 & $< 1e^{-300}$ \\
& Expert vs Actual  & -514.73 & $< 1e^{-300}$ \\
& ATE vs Expert     & 174.52  & $< 1e^{-300}$ \\
& LP vs Expert      & 177.90  & $< 1e^{-300}$ \\
& ATE vs LP         & -24.13  & $1.28 \times 10^{-128}$ \\

\midrule
\multicolumn{4}{l}{\textit{Cost}} \\
\midrule
& ATE vs Actual     & -159.44 & $< 1e^{-300}$ \\
& LP vs Actual      & 35.90   & $3.58 \times 10^{-282}$ \\
& Expert vs Actual  & -159.44 & $< 1e^{-300}$ \\
& ATE vs Expert     & -23.74  & $1.50 \times 10^{-124}$ \\
& LP vs Expert      & 172.24  & $< 1e^{-300}$ \\
& ATE vs LP         & -407.83 & $< 1e^{-300}$ \\

\bottomrule
\end{tabular}
\caption{US$>$AU: Paired t-test results comparing delivery time and cost across strategies using matched samples.}
\label{tab:ttest_results}
\end{table}

\begin{figure*}[t]
    \centering

    \includegraphics[width=0.85\textwidth]{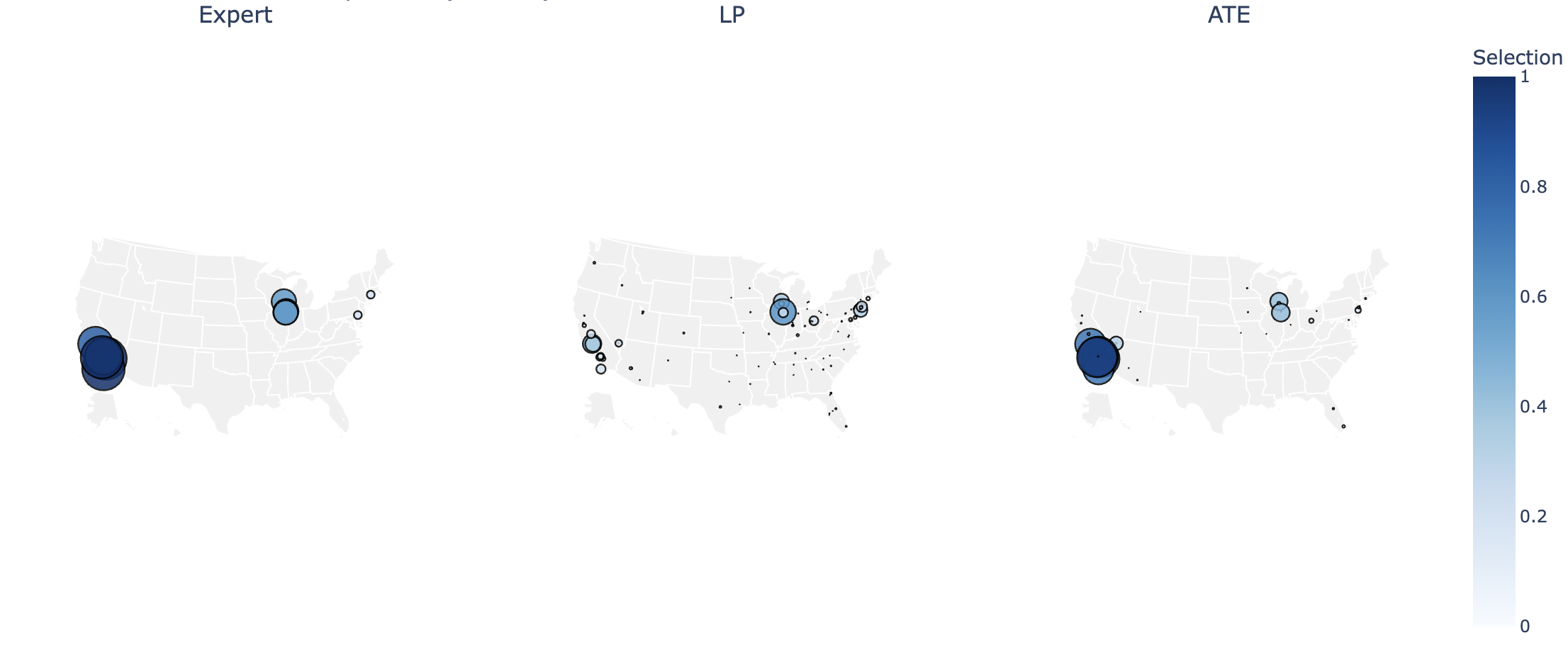}

    \vspace{0.3em}

    \includegraphics[width=0.85\textwidth]{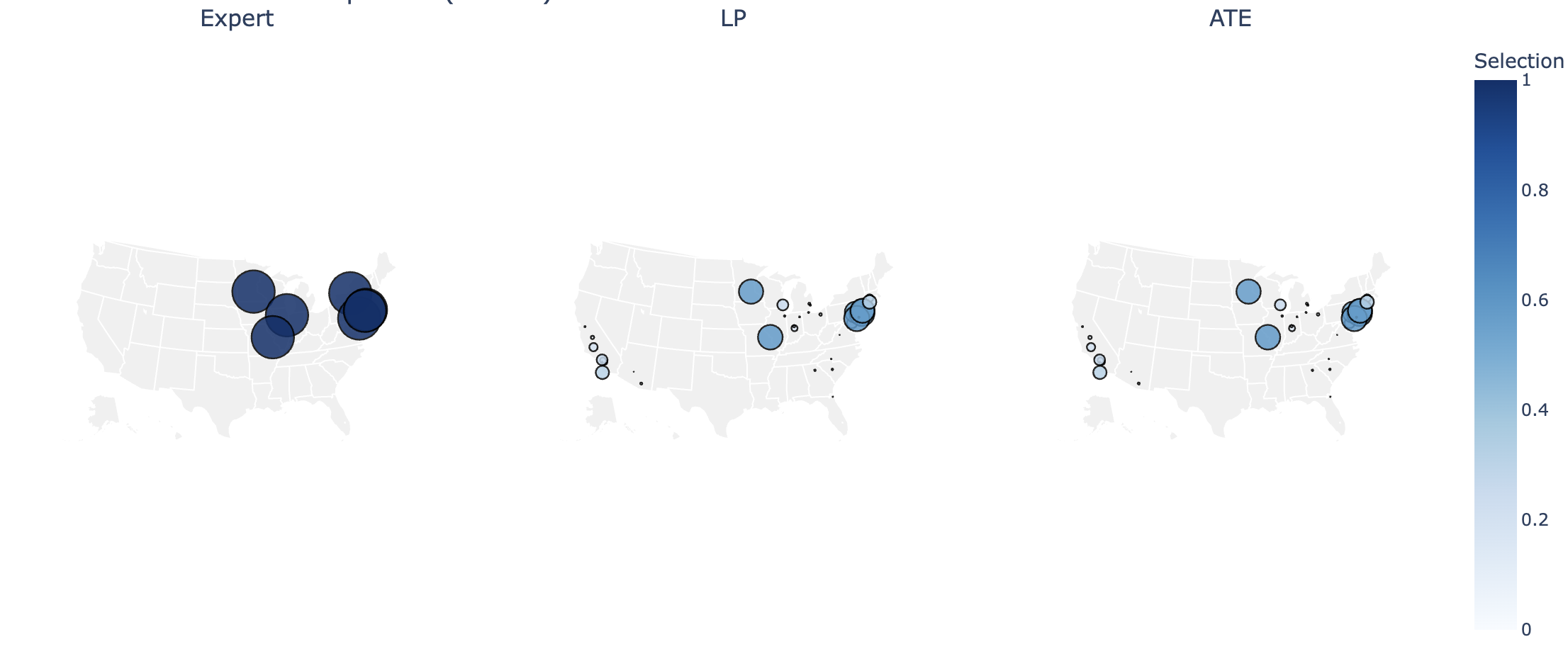}

    \vspace{0.3em}

    \includegraphics[width=0.85\textwidth]{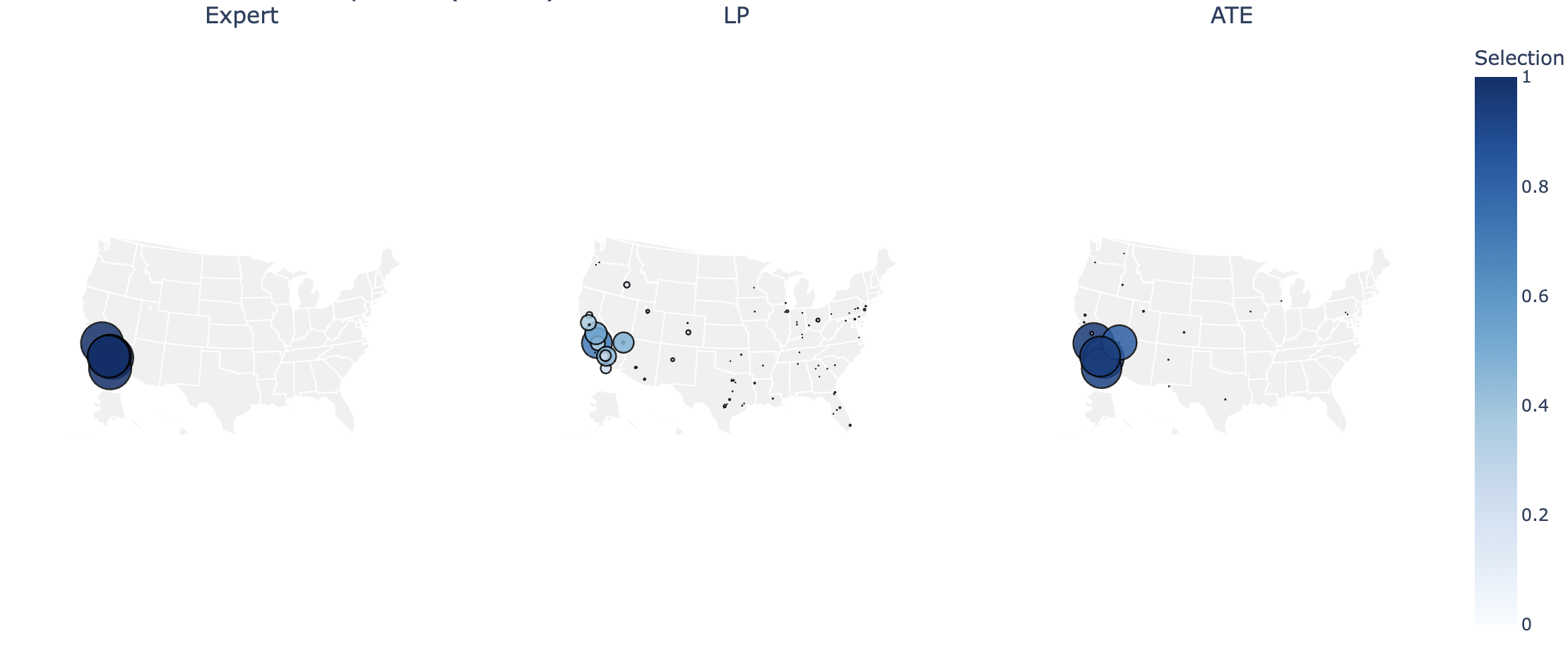}

    \caption{Spatial distribution of fulfillment center selection frequency under alternative policies. Each row shows the spatial distribution across U.S. warehouses for a distinct policy (Expert, LP, and XB FC Selection Model (ATE)). Marker size and color are proportional to normalized selection frequency, with larger and darker markers indicating higher FC utilization. The figure highlights how alternative policies redistribute fulfillment demand relative to the expert baseline.}
    \label{fig:three_graphs}
\end{figure*}

    
    
    
    
    

\begin{figure*}[htbp]
    \centering
    
    \begin{subfigure}[b]{0.32\textwidth}
        \centering
        \includegraphics[width=\textwidth]{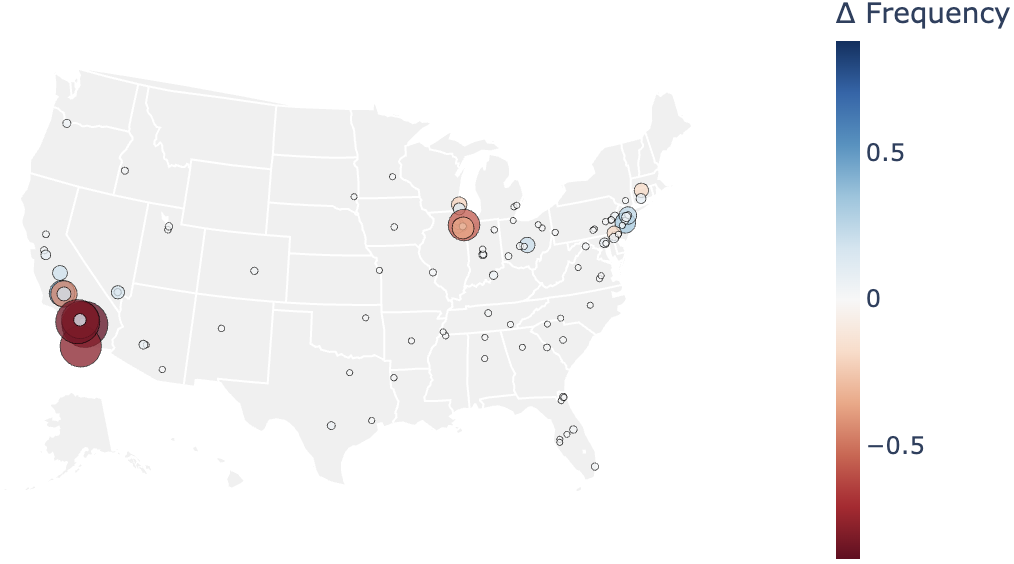}
        \caption{US$>$SG}
        \label{fig:graph1}
    \end{subfigure}
    \hfill
    \begin{subfigure}[b]{0.32\textwidth}
        \centering
        \includegraphics[width=\textwidth]{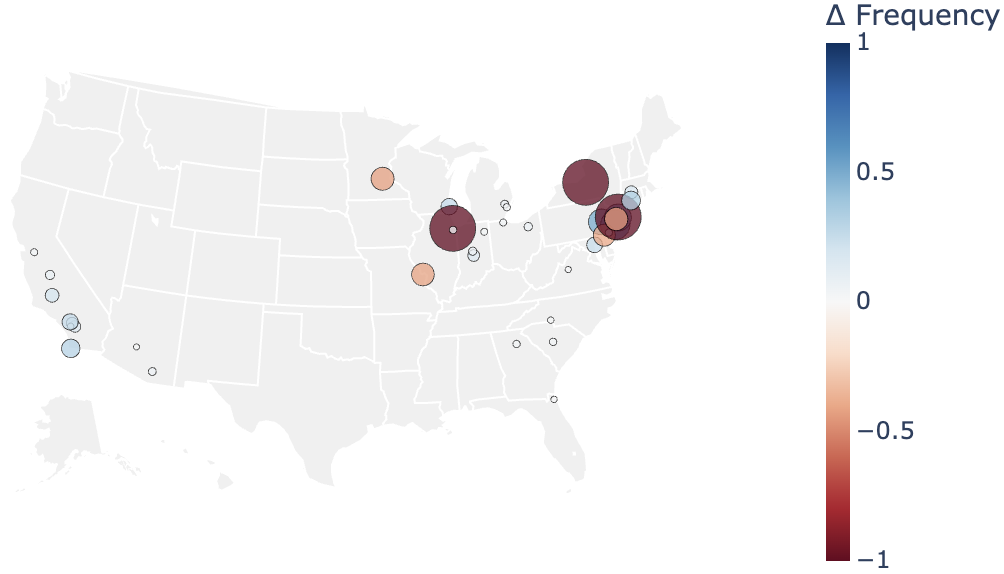}
        \caption{US$>$DE}
        \label{fig:graph2}
    \end{subfigure}
    \hfill
    \begin{subfigure}[b]{0.32\textwidth}
        \centering
        \includegraphics[width=\textwidth]{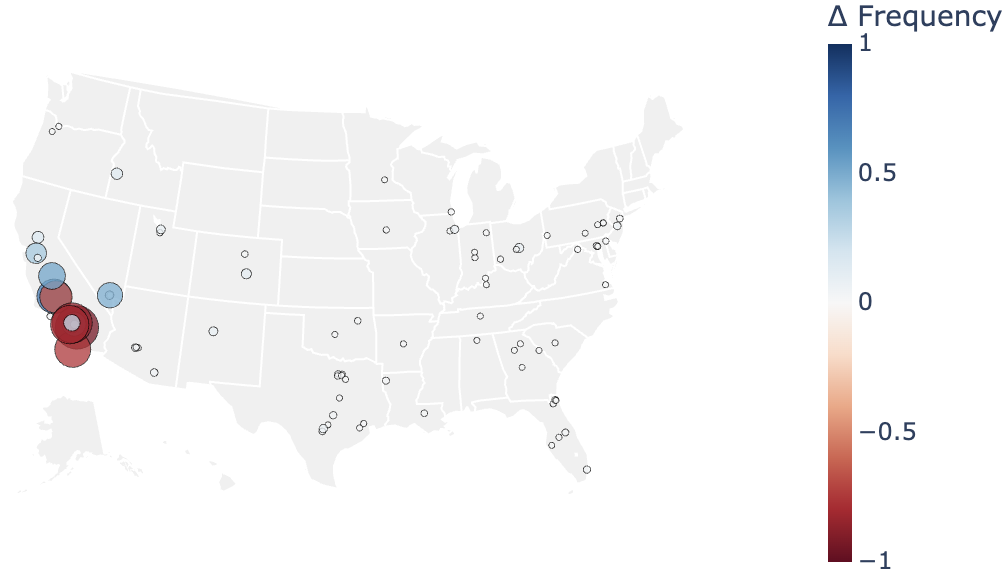}
        \caption{US$>$AU}
        \label{fig:graph3}
    \end{subfigure}
    
    \caption{Spatial Comparison of LP FC Selection Policies Relative to Expert Baseline
    This figure compares fulfillment center (FC) selection frequency differences relative to an expert-defined policy across U.S. warehouses. Each point represents a warehouse location, with marker size proportional to the magnitude of the deviation and color indicating the direction and magnitude of change in selection frequency (red = higher-than-expert selection, blue = lower-than-expert selection). The visualization highlights spatial heterogeneity in how alternative policies redistribute FC utilization compared to the baseline expert strategy.}
    \label{fig:three_graphs}
\end{figure*}

\clearpage
\FloatBarrier

\bibliographystyle{icml2026}
\bibliography{references}

@misc{deshpande2020,
  author = {Deshpande, V. and Pendem, P.},
  title  = {Logistics performance, ratings, and its impact on customer purchasing behavior and sales in e-commerce platforms},
  year   = {2020},
  note   = {Forthcoming}
}

@article{li2022,
  author  = {Li, Z.},
  title   = {The impact of logistics service quality of cross border e-commerce business on customer satisfaction},
  journal = {BCP Business \& Management},
  volume  = {29},
  pages   = {130--134},
  year    = {2022},
  doi     = {10.54691/bcpbm.v29i.2203}
}

@article{sinha2026,
  author  = {Sinha, A. and Agte, J. and Allgor, R. and Lara, C. L. and
             Agiwal, A. and Atakan, S. and Campo, L. and Cezik, T. and
             Chen, D. and Chen, Q. and Fischer, J. and Gor, E. and
             Kabbani, N. and Kennedy, R. and Krishnan, K. and Li, Y. and
             Mamadapur, S. M. and Mathur, T. and McCabe, N. and
             Mildebrath, D. and Powell, E. and Qualizza, A. and
             Schroeder, D. and Seth, N. and Si, X. and Sinha, K. and
             Stegner, D. and Xiao, J. and Zhang, L. and Zhang, S. and
             Zou, J.},
  title   = {Regionalize and Scale: Amazon's Fulfillment Network Design for Faster and Cheaper Delivery},
  journal = {INFORMS Journal on Applied Analytics},
  volume  = {56},
  number  = {1},
  pages   = {23--41},
  year    = {2026},
  doi     = {10.1287/inte.2025.0295}
}

@misc{oneill2023,
  author       = {O'Neill, S.},
  title        = {Sizing down to scale up: How Amazon reworked its fulfillment network to meet customer demand},
  year         = {2023},
  month        = jul,
  day          = {24},
  howpublished = {\url{https://www.amazon.science/news-and-features/how-amazon-reworked-its-fulfillment-network-to-meet-customer-demand}},
  note         = {Accessed: 2026-09-09}
}

@article{xu2009,
  author  = {Xu, P. J. and Allgor, R. and Graves, S. C.},
  title   = {Benefits of reevaluating real-time order fulfillment decisions},
  journal = {Manufacturing \& Service Operations Management},
  volume  = {11},
  number  = {2},
  pages   = {340--355},
  year    = {2009},
  doi     = {10.1287/msom.1080.0222}
}

@article{lara2023,
  author  = {Lara, C. L. and Koenemann, J. and Nie, Y. and de Souza, C. C.},
  title   = {Scalable timing-aware network design via Lagrangian decomposition},
  journal = {European Journal of Operational Research},
  volume  = {309},
  number  = {1},
  pages   = {152--169},
  year    = {2023},
  doi     = {10.1016/j.ejor.2023.01.018}
}

@misc{amazonscience2022,
  author       = {{Amazon Science}},
  title        = {Maximizing the efficiency of Amazon's own delivery networks},
  year         = {2022},
  howpublished = {\url{https://www.amazon.science/blog/maximizing-the-efficiency-of-amazons-own-delivery-networks}},
  note         = {Accessed: 2026-09-09}
}

@inproceedings{kassmann2006,
  author    = {Kassmann, D. and Allgor, R.},
  title     = {Supply chain design, management and optimization},
  booktitle = {16th European Symposium on Computer Aided Process Engineering and 9th International Symposium on Process Systems Engineering},
  editor    = {Marquardt, W. and Pantelides, C.},
  volume    = {21},
  pages     = {101--106},
  year      = {2006},
  publisher = {Elsevier B.V.},
  doi       = {10.1016/S1570-7946(06)80031-3}
}

@article{allgor2023,
  author  = {Allgor, R. and Cezik, T. and Chen, D.},
  title   = {Algorithm for robotic picking in Amazon fulfillment centers enables humans and robots to work together effectively},
  journal = {INFORMS Journal on Applied Analytics},
  volume  = {53},
  number  = {4},
  pages   = {266--282},
  year    = {2023},
  doi     = {10.1287/inte.2022.1143}
}

@article{cecil2024,
  author  = {Cecil, P.},
  title   = {Cross-border supply chain optimization: Strategies for managing international operations while maintaining speed and cost efficiency},
  journal = {International Journal of Scientific Research and Management},
  volume  = {12},
  number  = {5},
  pages   = {6565--6588},
  year    = {2024},
  doi     = {10.18535/ijsrm/v12i05.em23}
}

@inproceedings{guan2021,
  author    = {Guan, X. and Zhou, H. Y. and Zhang, L. Y. and Guo, X. M. and Xie, S. F.},
  title     = {Model design of cross-border e-commerce overseas warehouse location selection based on game theory},
  booktitle = {Proceedings of the 2021 International Conference on Bioinformatics and Intelligent Computing},
  pages     = {226--231},
  year      = {2021},
  doi       = {10.1145/3448748.3448784}
}

@article{acimovic2015,
  author  = {Acimovic, J. and Graves, S. C.},
  title   = {Making better fulfillment decisions on the fly in an online retail environment},
  journal = {Manufacturing \& Service Operations Management},
  volume  = {17},
  number  = {1},
  pages   = {34--51},
  year    = {2015},
  doi     = {10.1287/msom.2014.0505}
}

@incollection{jasin2019,
  author    = {Jasin, S. and Sinha, A. and Uichanco, J.},
  title     = {Omnichannel operations: Challenges, opportunities, and models},
  booktitle = {Operations in an Omnichannel World},
  editor    = {Gallino, S. and Moreno, A.},
  publisher = {Springer},
  address   = {Cham},
  pages     = {15--34},
  year      = {2019},
  doi       = {10.1007/978-3-030-20119-7_2}
}

\clearpage
\end{document}